\documentclass{article}

\usepackage{PRIMEarxiv}

\usepackage[utf8]{inputenc}
\usepackage[T1]{fontenc}
\usepackage{hyperref}
\usepackage{url}
\usepackage{booktabs}
\usepackage{amsfonts}
\usepackage{microtype}
\usepackage{graphicx}
\usepackage[numbers]{natbib}
\usepackage{hyperref}

\usepackage{amsmath,amssymb,amsthm}
\usepackage{stmaryrd}
\newtheorem{theorem}{Theorem}
\newtheorem{lemma}{Lemma}
\newtheorem{corollary}{Corollary}

\providecommand{\Description}[1]{}

\usepackage{booktabs}
\usepackage{tabularx}
\usepackage{enumitem}
\usepackage{url}
\makeatletter\g@addto@macro\UrlBreaks{\do\_\do\-}\makeatother
\usepackage{tikz}
\usetikzlibrary{positioning,arrows.meta,fit}
\tikzset{
  pics/contact/.style 2 args={code={\begin{scope}[xshift=0.3cm]
    \draw (-0.3,0) -- (-0.15,0) (0.15,0) -- (0.3,0);
    \draw[line width=0.6pt] (-0.15,-0.17) -- (-0.15,0.17) (0.15,-0.17) -- (0.15,0.17);
    \node[font=\scriptsize\sffamily, inner sep=0] at (0,0) {#2};
    \node[font=\scriptsize\ttfamily, above, inner sep=1pt] at (0,0.17) {#1};\end{scope}}},
  pics/ncontact/.style={code={\begin{scope}[xshift=0.3cm]
    \draw (-0.3,0) -- (-0.15,0) (0.15,0) -- (0.3,0);
    \draw[line width=0.6pt] (-0.15,-0.17) -- (-0.15,0.17) (0.15,-0.17) -- (0.15,0.17);
    \draw (-0.07,-0.13) -- (0.07,0.13);
    \node[font=\scriptsize\ttfamily, above, inner sep=1pt] at (0,0.17) {#1};\end{scope}}},
  pics/coil/.style 2 args={code={\begin{scope}[xshift=0.3cm]
    \draw (-0.3,0) -- (-0.188,0) (0.188,0) -- (0.3,0);
    \draw (-0.09,0.17) arc[start angle=120, end angle=240, radius=0.196];
    \draw (0.09,0.17) arc[start angle=60, end angle=-60, radius=0.196];
    \node[font=\scriptsize\sffamily, inner sep=0] at (0,0) {#2};
    \node[font=\scriptsize\ttfamily, above, inner sep=1pt] at (0,0.17) {#1};\end{scope}}},
  rail/.style={line width=1.2pt},
}
\usepackage{listings}
\usepackage[capitalise,noabbrev]{cleveref}
\crefname{equation}{Equation}{Equations}
\newcommand{\fitwidth}[1]{\resizebox{\ifdim\width>\linewidth\linewidth\else\width\fi}{!}{#1}}
\newcommand{\iss}[1]{\##1}
\newcommand{\anonhash}[1]{\texttt{#1}}

\usepackage[acronym,nomain,nonumberlist]{glossaries}
\setacronymstyle{long-short}
\glsdisablehyper

\newacronym{plc}{PLC}{programmable logic controller}
\newacronym{ld}{LD}{ladder diagram}
\newacronym{smt}{SMT}{satisfiability modulo theories}
\newacronym{bmc}{BMC}{bounded model checking}
\newacronym{fbd}{FBD}{function block diagram}
\newacronym{sfc}{SFC}{sequential function chart}
\newacronym{ssa}{SSA}{static single assignment}
\newacronym{esbmc}{ESBMC}{efficient SMT-based context-bounded model checker}
\newacronym{scl}{SCL}{Structured Control Language}
\newacronym{stl}{STL}{Statement List}
\newacronym{svcomp}{SV-COMP}{Competition on Software Verification}

\usepackage{xcolor}

\title{ESBMC-PLC: Verifying IEC~61131-3 Ladder Diagrams with a Validated Translation to SMT-Based Model Checking}

\author{
  Pierre Dantas \\
  Department of Computer Science, The University of Manchester \\
  Department of Electrical Engineering, Federal University of Amazonas (UFAM) \\
  \texttt{pierre.dantas@manchester.ac.uk} \\
  \And
  Lucas Cordeiro \\
  Department of Computer Science, The University of Manchester \\
  Department of Electrical Engineering, Federal University of Amazonas (UFAM) \\
  \texttt{lucas.cordeiro@manchester.ac.uk} \\
  \And
  Waldir Junior \\
  Department of Electrical Engineering, Federal University of Amazonas (UFAM) \\
  Manaus, AM, Brazil \\
  \texttt{waldirjr@ufam.edu.br} \\
}

\begin{document}
\maketitle
\glsresetall

\begin{abstract}
\Gls{ld} is the relay-circuit language of IEC~61131-3 in which many \gls{plc} safety interlocks are written, and
model checkers built on \gls{smt} solvers do not read it. A verdict reached through a translation is only as good as
that translation. We present ESBMC-PLC, which verifies \gls{ld} programs with the \gls{esbmc}, and argue
that its translation is faithful. It encodes the scan cycle as one loop with nondeterministic inputs,
resolves graphical PLCopen XML networks into ordered rungs, models timers in milliseconds, and refuses programs outside
an explicit subset. We prove, under eight stated hypotheses, that one iteration of the emitted loop agrees with one scan of a semantics
for the subset, so scan-count verdicts on programs without user function blocks hold on that semantics. For timing-mode
verdicts, a refinement theorem for MATIEC's timers gives evidence, not proof. One \gls{smt} query per program also
compares the emitted loop with the C that MATIEC generates. On the 241 labeled \gls{ld} variants of a public benchmark
suite, ESBMC-PLC returns no wrong verdict against the current labels and proves 127 variants safe; \gls{esbmc} on
MATIEC's C for the same programs proves 2. PLCverif with NuSMV, which shares neither our frontend nor \gls{esbmc}, agrees
with ESBMC-PLC on all 207 of the 208 variants within the theorem's scope that it can run. On 413 \gls{ld} programs from public GitHub
repositories, none of the 104 that the per-program query can check differs from MATIEC. Every $k$-induction proof closes at
$k = 2$. \gls{esbmc}'s inductive step proves only properties that one scan preserves, so properties that depend on a longer
history stay unproved.

\end{abstract}

\keywords{programmable logic controllers, IEC~61131-3, ladder diagram, bounded model checking, $k$-induction,
translation validation, formal semantics, timing abstraction}

\glsresetall

\section{Introduction}
\label{sec:intro}

Interlocks, emergency stops, and burner sequences on industrial plants run as programs on \glspl{plc}, and many of
those programs are written in \gls{ld}, the relay-circuit notation of
IEC~61131-3~\cite{iec61131-3-2025}. An \gls{ld} program looks simple: rungs of contacts and coils, a few timers. Its
behavior is not, because it runs as a scan that never ends, timers and edge contacts carry state from one scan to the
next, and the order in which coils fire decides what the outputs hold when the scan ends. The running example of this
paper, a stairwell-light controller of two rungs (\cref{sec:example}), has a manual toggle that can never turn the
light on, and nothing in the drawing shows it.

Model checkers built on \gls{smt} solvers can explore every input sequence of such a program, but they read C or their own modeling
languages, not \gls{ld}. Verification therefore goes through a translation, and the verdict is only as good as that
translation: a frontend that misreads a contact, drops a timer, or picks the wrong coil order verifies a program the
engineer did not write, and returns a verdict with nothing to show the difference. Graphical PLCopen XML, the form that
open editors export, adds a step of its own, since the file gives no rung order and the translator has to reconstruct
the one the runtime uses.

A frontend of its own also changes what can be proved. The \gls{esbmc} already reads the C that the open-source MATIEC compiler
generates from a PLCopen file, so an engineer could verify that C instead. Both routes find nearly the same violations.
They part on proofs. On the 127 benchmark variants that ESBMC-PLC proves safe, the C route proves 2, times out or ends
UNKNOWN on 124, and has no C for one, whereas k-induction on the scan loop that ESBMC-PLC emits closes on all 127 at
$k = 2$ (\cref{sec:rq-c}). Facts about MATIEC's variable records, written by hand, give most of these proofs back to
the C route, but not those of five timer programs.

This paper presents ESBMC-PLC, the \gls{ld} frontend of the \gls{esbmc} model checker~\cite{Gadelha2018esbmc,Menezes2024esbmc},
together with the argument that makes its verdicts trustworthy: that its translation is faithful. Its contributions are
four.
\begin{enumerate}[leftmargin=*]
\item \textbf{The approach} (\cref{sec:approach}). ESBMC-PLC accepts an explicit subset of textual and graphical \gls{ld} and
  refuses the rest with a diagnostic, apart from one known gap that Appendix~\ref{app:gaps} describes. It encodes the scan cycle as one GOTO loop with nondeterministic inputs,
  translates rungs and resolves graphical networks into them, models timers with a duration per scan in milliseconds,
  and compiles five property kinds into monitors that \gls{esbmc} checks with bounded model checking and k-induction.
\item \textbf{A correctness argument} (\cref{sec:correctness}). We give the subset a scan semantics and prove that one
  iteration of the emitted loop and one scan of the semantics agree, under eight stated hypotheses. Under three
  preconditions on the runtime's clock and presets, MATIEC's timers refine the timing model, and negative controls show that a real
  run exceeds a proved bound when a precondition fails.
\item \textbf{Per-program translation validation} (\cref{sec:tv}). One \gls{smt} query per program compares a scan of the
  emitted GOTO program with a scan of the C that MATIEC generates from the same file, with a mutation control that
  rules out a harness unable to see a difference.
\item \textbf{An evaluation} (\cref{sec:evaluation}) on the 241 \gls{ld} variants of a public benchmark suite: 127 variants
  proved safe where \gls{esbmc} on MATIEC's C of the same programs proves 2, exact single-timer deadlines proved by
  k-induction at $k = 2$ for presets up to 20~s, no wrong verdict against the current labels, 17 programs outside the subset refused, and 210 of
  229 variants equivalent to MATIEC after one scan with every difference traced to one cause. PLCverif with NuSMV agrees
  with ESBMC-PLC on all 207 of the 208 variants within the theorem's scope that it can run (Beremiz produces no Structured
  Text for the other). Two parametric families measure properties that
  depend on history: ESBMC-PLC finds their bugs but, like \gls{esbmc}'s k-induction in general, proves none of their safe
  variants. A second corpus of 413 programs from 138 public GitHub repositories, none from the developers of the tool or the suite, tests the
  translation: the subset accepts 115, and the 104 that the check could run all agree with
  MATIEC.
\end{enumerate}
Upstream \gls{esbmc} contains the frontend and its subset checks. It also fixes three departures that the evaluated build
keeps (Appendix~\ref{app:departures}). The timing mode, the block and monitor invariants, and
translation validation are on a public fork of \gls{esbmc} and not yet merged upstream; the artifact pins that build and
holds the suite run and the scripts behind every number~\cite{esbmcplc-artifact}. \Cref{sec:related} discusses
earlier arXiv preprints on ESBMC-PLC.

\section{Background and running example}
\label{sec:background}

\subsection{Ladder diagrams and the scan cycle}
\label{sec:ld}

A \gls{plc} runs its program in a loop. Each iteration, a \emph{scan}, samples the inputs,
runs the program once from top to bottom, and leaves the outputs for the hardware to apply. \Gls{ld}, one of
the languages of IEC~61131-3~\cite{iec61131-3-2025}, writes the program as rungs between two vertical power rails. A
\emph{contact} on a variable passes power when the variable is true (\texttt{-[ ]-}), when it is false
(\texttt{-[/]-}), or on the scan where it rises or falls (\texttt{-[P]-}, \texttt{-[N]-}). A \emph{coil} at the right
end of a rung writes the power it receives to a variable; a set coil (\texttt{-(S)-}) writes true when powered, and a
reset coil (\texttt{-(R)-}) writes false. Contacts in series form a conjunction, branches in parallel a disjunction.
Function blocks sit on rungs as boxes with named pins: on-delay, off-delay, and pulse timers (TON, TOF, TP),
up and down counters (CTU, CTD), and arithmetic.
Timers and counters keep state from one scan to the next, so an \gls{ld} program is a reactive system, not a Boolean
formula. \Cref{fig:ld-elements} draws each of these elements.

\begin{figure}[!ht]
\centering
\begin{tikzpicture}[x=1cm, y=1cm]
  \draw[rail] (0,0.45) -- (0,-3.35);
  \draw[rail] (8,0.45) -- (8,-3.35);
  % Q := (A or rising C) and not B
  \draw (0,0) -- (0.9,0) pic {contact={A}{}} (1.5,0) -- (3.6,0) pic {ncontact={B}} (4.2,0) -- (6.4,0) pic {coil={Q}{}} (7,0) -- (8,0);
  \draw (0,-0.8) -- (0.9,-0.8) pic {contact={C}{P}} (1.5,-0.8) -- (2,-0.8) -- (2,0);
  % a TON on D sets M; a falling edge of E resets it
  \draw (0,-1.9) -- (0.9,-1.9) pic {contact={D}{}} (1.5,-1.9) -- (2.8,-1.9);
  \draw (2.8,-1.6) rectangle (4.4,-2.75);
  \node[font=\scriptsize\sffamily, above] at (3.6,-1.6) {T1};
  \node[font=\scriptsize\sffamily] at (3.6,-1.72) {TON};
  \node[font=\tiny\sffamily, right] at (2.8,-1.9) {IN};
  \node[font=\tiny\sffamily, right] at (2.8,-2.45) {PT};
  \node[font=\tiny\sffamily, left] at (4.4,-1.9) {Q};
  \node[font=\tiny\sffamily, left] at (4.4,-2.45) {ET};
  \draw (2.8,-2.45) -- (2.5,-2.45) node[left, font=\scriptsize\ttfamily, inner sep=1pt] {T\#5s};
  \draw (4.4,-1.9) -- (6.4,-1.9) pic {coil={M}{S}} (7,-1.9) -- (8,-1.9);
  \draw (0,-3.2) -- (0.9,-3.2) pic {contact={E}{N}} (1.5,-3.2) -- (6.4,-3.2) pic {coil={M}{R}} (7,-3.2) -- (8,-3.2);
\end{tikzpicture}
\caption{Three rungs between the power rails. The first writes $Q := (A \vee \uparrow C) \wedge \neg B$, where
$\uparrow C$ holds on the scan where $C$ rises. In the second, timer \texttt{T1} sets $M$ once $D$ has been true for
5~s. In the third, a falling edge of $E$ resets $M$.}
\label{fig:ld-elements}
\Description{A ladder diagram of three rungs between two vertical power rails, with contacts, a TON block, and normal, set, and reset coils.}
\end{figure}

Engineering tools exchange \gls{ld} programs as PLCopen XML~\cite{plcopen-tc6}. A file describes a rung either textually, as
an ordered list of elements, or graphically, as boxes with coordinates and wires between their connection points.
Beremiz and the OpenPLC Editor export the graphical form. It carries no rung order: the runtime derives one from the
drawing, and a verifier has to derive the same one.

\subsection{\texorpdfstring{\acrshort{esbmc}}{ESBMC}}
\label{sec:esbmc}

\gls{esbmc}~\cite{Gadelha2018esbmc,Menezes2024esbmc} is a bounded model checker built on \gls{smt} solvers. Each
frontend lowers its language to one intermediate representation, the GOTO program, and everything after that point
is shared: symbolic execution into \gls{ssa}, bit-precise encoding of integer and floating-point arithmetic, and a choice of
\gls{smt} solver. Two of its engines matter here. \Gls{bmc} unrolls loops up to a bound and returns a
counterexample when a property fails within it. k-induction adds an inductive step over the same encoding and, when
the step closes, proves the property for every number of loop iterations. A scan loop never terminates, so among
\gls{esbmc}'s engines only k-induction can prove that a property holds forever; \gls{bmc} finds the violations.

\subsection{Running example}
\label{sec:example}

We follow one program through the paper: the stairwell-light controller that ships with the CONTROLLINO \gls{plc} board, a
graphical PLCopen file with five variables and one timer, run by a 20~ms task. Two rungs carry the logic. A rising edge
of either push button, \texttt{control\_button\_up} or \texttt{control\_button\_down}, is meant to toggle
\texttt{lights\_buttons\_state}: a set coil fires when the variable is false, a reset coil when it is true. When the
motion sensor \texttt{stairs\_pir\_sensor} rises while the toggle is off, it drives the input of \texttt{TOF0}, an
off-delay timer with a preset of 20~s, and \texttt{stairs\_light} is \texttt{TOF0.Q} or the toggle.
\Cref{fig:stairs} shows both rungs as the file draws them.

\begin{figure}[t]
\centering
\fitwidth{\begin{tikzpicture}[x=1cm, y=1cm]
  \draw[rail] (0,0.45) -- (0,-3.65);
  \draw[rail] (12,0.45) -- (12,-3.65);
  % rung 1: a rising edge of either button sets or resets the toggle
  \draw (0,0) -- (1.3,0) pic {contact={control\_button\_up}{P}} (1.9,0) -- (5.7,0) pic {ncontact={lights\_buttons\_state}} (6.3,0) -- (9.7,0) pic {coil={lights\_buttons\_state}{S}} (10.3,0) -- (12,0);
  \draw (0,-0.9) -- (1.3,-0.9) pic {contact={control\_button\_down}{P}} (1.9,-0.9) -- (5.7,-0.9) pic {contact={lights\_buttons\_state}{}} (6.3,-0.9) -- (9.7,-0.9) pic {coil={lights\_buttons\_state}{R}} (10.3,-0.9) -- (12,-0.9);
  \draw (3,0) -- (3,-0.9);
  \fill (3,0) circle (1.2pt) (3,-0.9) circle (1.2pt);
  % rung 2: the sensor drives TOF0; the light is TOF0.Q or the toggle
  \draw (0,-2.2) -- (1.3,-2.2) pic {contact={stairs\_pir\_sensor}{P}} (1.9,-2.2) -- (4.2,-2.2) pic {ncontact={lights\_buttons\_state}} (4.8,-2.2) -- (5.9,-2.2);
  \draw (5.9,-1.9) rectangle (7.5,-3.05);
  \node[font=\scriptsize\sffamily, above] at (6.7,-1.9) {TOF0};
  \node[font=\scriptsize\sffamily] at (6.7,-2.02) {TOF};
  \node[font=\tiny\sffamily, right] at (5.9,-2.2) {IN};
  \node[font=\tiny\sffamily, right] at (5.9,-2.75) {PT};
  \node[font=\tiny\sffamily, left] at (7.5,-2.2) {Q};
  \node[font=\tiny\sffamily, left] at (7.5,-2.75) {ET};
  \draw (5.9,-2.75) -- (5.6,-2.75) node[left, font=\scriptsize\ttfamily, inner sep=1pt] {T\#20s};
  \draw (7.5,-2.2) -- (9.7,-2.2) pic {coil={stairs\_light}{}} (10.3,-2.2) -- (12,-2.2);
  \draw (0,-3.45) -- (1.3,-3.45) pic {contact={lights\_buttons\_state}{}} (1.9,-3.45) -- (8.4,-3.45) -- (8.4,-2.2);
  \fill (8.4,-2.2) circle (1.2pt);
\end{tikzpicture}}
\caption{The two rungs of the running example. A rising edge of either button sets the toggle when it is false
and resets it when it is true; a rising edge of the motion sensor, with the toggle off, starts the off-delay timer
\texttt{TOF0}, and the light is \texttt{TOF0.Q} or the toggle.}
\label{fig:stairs}
\Description{The stairwell ladder diagram: two button contacts in parallel feed a set and a reset coil on the toggle; the sensor contact and a negated toggle contact feed an off-delay timer whose output, in parallel with the toggle, drives the light coil.}
\end{figure}

The off-delay keeps the light on for 20~s after the sensor's pulse, which is the purpose of a stairwell light. An
engineer's first attempt at a property says otherwise:
\begin{equation}
\label{eq:p1}
\texttt{stairs\_light} \Rightarrow \texttt{stairs\_pir\_sensor} \vee \texttt{lights\_buttons\_state}.
\end{equation}
ESBMC-PLC refutes \cref{eq:p1} in the base case of k-induction at $k = 2$ (\cref{tab:stairs-cex}): one scan after the
pulse, the timer still holds the light and the sensor is low.

\begin{table}[t]
\centering
\small
\caption{The counterexample ESBMC-PLC returns for \cref{eq:p1}, as values at the end of each scan. Both buttons rise
in scan 1, and \texttt{lights\_buttons\_state} stays false.}
\label{tab:stairs-cex}
\begin{tabular}{@{}cccccc@{}}
\toprule
Scan & \texttt{stairs\_pir\_sensor} & \texttt{control\_button\_up}, \texttt{\_down} & \texttt{TOF0.IN} &
\texttt{lights\_buttons\_state} & \texttt{stairs\_light} \\
\midrule
1 & 1 & 1, 1 & 1 & 0 & 1 \\
2 & 0 & 0, 0 & 0 & 0 & 1 \\
\bottomrule
\end{tabular}
% esbmc-p2 5efb80de72: regression/ld/stairs_light_hold_fail, --k-induction --unlimited-k-steps
\end{table}
 It also warns that the toggle has a set and a reset coil that can fire in the
same scan. They do. On the scan a button rises, the set coil fires first, the reset coil reads the value just written
and clears it, and so \texttt{lights\_buttons\_state} is false at every scan boundary: the manual toggle never turns the
light on. \Cref{sec:approach} shows how the program becomes a GOTO loop, and \cref{sec:deadlines} turns
\cref{eq:p1} into the question an engineer would ask, namely how long the light can stay on after the last motion.

\section{The ESBMC-PLC approach}
\label{sec:approach}

ESBMC-PLC turns an \gls{ld} program and a property file into a GOTO program that \gls{esbmc} checks with its existing engines.
This section describes each step of that translation; \cref{sec:correctness} argues that the result means what
the diagram means.

\subsection{Overview}
\label{sec:overview}

\Cref{fig:pipeline} shows the pipeline. A user runs \texttt{esbmc prog.ld -{}-ld-props props.yaml -{}-k-induction}. The
frontend parses the PLCopen file, checks types, and either refuses the program with a diagnostic or continues;
graphical networks are first resolved into rungs. From the rungs, the translator emits one GOTO function,
\texttt{ld::scan\_loop}, a \texttt{while (true)} loop whose body is one scan, and appends a monitor for each property
of the YAML file. \gls{esbmc} then runs unchanged. The same option set reads the C that the MATIEC
compiler~\cite{matiec} produces from the same PLCopen file, which \cref{sec:tv} uses to check the
translation of each program.

\begin{figure}[t]
\centering
\fitwidth{\begin{tikzpicture}[
  box/.style={draw, rounded corners=2pt, align=center, font=\footnotesize, minimum height=0.9cm, inner sep=3pt},
  verdict/.style={font=\footnotesize\bfseries},
  arr/.style={-{Stealth[length=2mm]}},
  node distance=0.45cm]
\node[box] (xml) {PLCopen XML\\textual or graphical};
\node[box, right=of xml] (parse) {parse, type check,\\refuse outside subset};
\node[box, right=of parse] (graph) {resolve graphical\\networks into rungs};
\node[box, right=of graph] (trans) {emit \texttt{ld::scan\_loop}\\and monitors};
\node[box, right=of trans] (esbmc) {\gls{esbmc}: \gls{bmc}\\or k-induction};
\node[box, above=0.35cm of trans] (yaml) {YAML properties};
\node[verdict, below=0.45cm of parse] (ref) {REFUSED};
\node[verdict, right=0.35cm of esbmc, align=left] (v) {SAFE\\VIOLATION\\UNKNOWN};
\node[box, below=1.1cm of graph] (matiec) {Beremiz and MATIEC:\\C of the same file};
\node[box, below=1.1cm of trans] (tv) {one-scan \gls{smt} query\\(\path{--ld-tv-check})};
\node[verdict, right=0.35cm of tv, align=left] (tvv) {equivalent\\different};
\draw[arr] (xml) -- (parse);
\draw[arr] (parse) -- (graph);
\draw[arr] (graph) -- (trans);
\draw[arr] (trans) -- (esbmc);
\draw[arr] (yaml) -- (trans);
\draw[arr] (parse) -- (ref);
\draw[arr] (esbmc) -- (v);
\draw[arr] (xml.south) |- (matiec.west);
\draw[arr] (matiec) -- (tv);
\draw[arr] (trans) -- (tv);
\draw[arr] (tv) -- (tvv);
\end{tikzpicture}}
\caption{The ESBMC-PLC pipeline. A PLCopen XML file and a YAML property file go in; the frontend refuses a program
outside the subset or emits a scan loop with one monitor per property, and \gls{esbmc} checks it unchanged, as it would a C
program. The lower branch is translation validation (\cref{sec:tv}), which compares one scan of the emitted loop with
the C that MATIEC generates from the same file.}
\Description{A left-to-right flow: PLCopen XML enters a parsing and type-checking stage that can end in REFUSED; the
rest flows to the graph resolver, then to the translator, which also receives YAML properties and emits the scan loop
with monitors; \gls{esbmc} checks it with \gls{bmc} or k-induction and returns SAFE, VIOLATION, or UNKNOWN. A lower branch sends the same
XML through Beremiz and MATIEC to C, and a one-scan \gls{smt} query compares that C with the emitted loop, answering
equivalent or different.}
\label{fig:pipeline}
\end{figure}

\subsection{The supported subset}
\label{sec:subset}

A program is one program organization unit run by one periodic task. Its body is a sequence of rungs built from the
elements of \cref{fig:grammar}, given textually or as a graphical network. The frontend refuses the rest
before translation and names the construct in its diagnostic: other body languages such as \gls{fbd} and \gls{sfc}, several tasks, wire
cycles, unknown block types and pin names, negated or transition-sensing coils, a wired enable input or output (\texttt{EN}, \texttt{ENO}) on
a standard block, comparison blocks in power flow, and a list of declaration conflicts given in
Appendix~\ref{app:semantics}. One gap remains: a jump placed beside textual rungs, which our build
and upstream \gls{esbmc} drop without a refusal (Appendix~\ref{app:gaps}). The frontend also accepts calls to user function blocks, whose Structured Text bodies a
separate component translates, but the correctness argument of \cref{sec:correctness} excludes them (H6), and the
evaluation reports them apart. Refusing is part of the method. A construct the semantics does not describe is never
translated approximately, because an approximate translation verifies a different program from the one in the
drawing and gives no sign of it.

\begin{figure}[t]
\centering
\fitwidth{\begin{tabular}{@{}r@{\;}c@{\;}l@{}}
contact & $::=$ & \texttt{-[ ]-} $v$ $\mid$ \texttt{-[/]-} $v$ $\mid$ \texttt{-[P]-} $v$ $\mid$ \texttt{-[N]-} $v$ \\
coil    & $::=$ & \texttt{-( )-} $v$ $\mid$ \texttt{-(S)-} $v$ $\mid$ \texttt{-(R)-} $v$ \\
block   & $::=$ & $T(\mathit{IN}, \mathit{PT}, Q, \mathit{ET})$ for $T \in \{\mathrm{TON}, \mathrm{TOF}, \mathrm{TP}\}$ \\
        & $\mid$ & $\mathrm{CTU}(\mathit{CU}, R, \mathit{PV}, Q, \mathit{CV})$ $\mid$ $\mathrm{CTD}(\mathit{CD}, \mathit{LD}, \mathit{PV}, Q, \mathit{CV})$ \\
        & $\mid$ & $\mathit{op}(\mathit{IN1}, \mathit{IN2}, \mathit{OUT})$ for $\mathit{op} \in \{\mathrm{ADD}, \mathrm{SUB}, \mathrm{MUL}, \mathrm{DIV}\}$ $\mid$ $\mathrm{MOVE}(\mathit{IN1}, \mathit{OUT})$
\end{tabular}}
\caption{Grammar of the supported \gls{ld} subset. $v$ ranges over declared variables of type BOOL, INT (16 bits), DINT, TIME (32 bits),
or REAL; the operands and result of an arithmetic block or MOVE are INT or DINT. Calls to user function blocks are
also accepted, but \cref{thm:scan} excludes them (H6).
Pins:
\textit{IN} input, \textit{PT} preset time, \textit{ET} elapsed time, \textit{Q} output, \textit{CU} and
\textit{CD} count inputs, \textit{R} reset, \textit{LD} load (not ladder diagram), \textit{PV} preset value,
\textit{CV} current value, \textit{IN1}, \textit{IN2}, and \textit{OUT} operands and result. DINT is a
32-bit integer and REAL a floating-point number.}
\label{fig:grammar}
\Description{A grammar in BNF of the supported ladder-diagram elements: contacts, coils, timer and counter blocks, arithmetic blocks, and MOVE.}
\end{figure}

\subsection{The scan-cycle encoding}
\label{sec:scan-encoding}

The loop body follows the read, execute, and latch structure that MATIEC and OpenPLC implement:
\begin{equation}
\label{eq:tscan}
T_{\mathit{scan}}(P) \;=\; \mathit{RI} \,;\; T_{\mathit{rung}}(R_1) \,;\; \dots \,;\; T_{\mathit{rung}}(R_n) \,;\;
\mathit{LATCH} \,;\; \mathit{MON}.
\end{equation}
$\mathit{RI}$ gives every input a nondeterministic value of its type, so a proof covers every input sequence. Rungs
run in order. $\mathit{LATCH}$ copies each variable that an edge contact senses into a shadow variable, and
$\mathit{MON}$ holds the property monitors, which therefore observe the store at the scan boundary.

Which variables $\mathit{RI}$ assigns decides what a verdict covers. The frontend reads it from the program's
interface section first (\texttt{VAR\_INPUT}), then from the address (\texttt{\%I}), and only then from use: a variable that the
program reads and never writes is treated as free. This use rule over-approximates, so it can raise a spurious alarm
but cannot hide a violation. A variable that another task or the environment may write, such as a global, is also
assigned by $\mathit{RI}$ on every scan unless the user asserts with \path{--ld-closed-world} that the program owns
it. Every other variable keeps its value across scans, which models the output image and internal markers. Types map
to bit-vectors of their IEC width: BOOL to a Boolean, INT to 16 bits, and DINT and TIME to 32 bits, so arithmetic wraps
where the controller's does; REAL maps to a double (Appendix~\ref{app:departures}).

\subsection{Translating rungs}
\label{sec:translation}

$T_{\mathit{rung}}$ walks a rung from left to right and keeps an accumulator expression $E$, which starts as true.
\Cref{tab:translation} lists what each element emits. Contacts emit no code; they extend $E$. A coil assigns $E$ at the
point where the coil stands, so every contact of the rung is read in the store current at the coil. The choice matters
when a rung resets a variable and reads it again further right; \cref{sec:semantics} shows that MATIEC makes the
same choice. A set coil assigns true when $E$ holds and a reset coil false, and two coils on one variable keep the
order of their rungs, so the later one wins in a scan where both are powered. The frontend warns when that can
happen, as it does in the running example.

\begin{table}[t]
\centering
\caption{What $T_{\mathit{rung}}$ emits per element. $E$ is the rung's accumulator, $\mathit{prev}_i$ the shadow of
block $i$'s input.}
\label{tab:translation}
\small
\begin{tabularx}{\linewidth}{@{}l>{\raggedright\arraybackslash}X@{}}
\toprule
Element & Emitted code \\
\midrule
contact on $v$ & none; $E := E \wedge \mathit{cv}$, with $\mathit{cv}$ the contact's test ($v$, $\neg v$, or an edge test
  against the shadow of $v$) \\
coil, set coil, reset coil & $v := \mathit{cast}_v(E)$; \texttt{if} $E$ \texttt{then} $v := \mathit{cast}_v(\mathit{true})$;
  \texttt{if} $E$ \texttt{then} $v := \mathit{cast}_v(\mathit{false})$ \\
TON, TOF, TP & read \textit{IN} (set to $E$ by a generated rung in a graphical network); start, hold, or expire the
  interval; increment \textit{ET} guarded by $\mathit{ET} < \mathit{PT}$; set $Q$; $\mathit{prev}_i := \mathit{IN}$ \\
CTU, CTD & CTU: count on a rising edge of \textit{CU} with a guard at \textit{PV}, then reset; CTD: load, else count
  down on a rising edge of \textit{CD} with a guard at 0; set $Q$; $\mathit{prev}_i$ \\
arithmetic, MOVE & $\mathit{OUT} := \mathit{cast}(\mathit{IN1} \;\mathit{op}\; \mathit{IN2})$, wrapping in the operands'
  common type \\
\bottomrule
\end{tabularx}
\end{table}

\Cref{lst:tof} shows the translation of the timer rung of the running example. The first two blocks compute the
power flow into the timer, a rising edge of the sensor while the toggle is off; the rest is the TOF. On the scan
where \textit{IN} falls the shadow still holds the old input, so that scan counts no elapsed time, as in MATIEC.

\begin{figure}[t]
\begin{lstlisting}
    pf9 := 0;  IF !(stairs_pir_sensor && !__edge_prev_stairs_pir_sensor) GOTO L1;  pf9 := 1;
L1: pf12 := 0;  IF !(pf9 && !lights_buttons_state) GOTO L2;  pf12 := 1;
L2: TOF0__IN := pf12;
    IF !TOF0__IN GOTO L3;  TOF0__ET := 0;  TOF0__Q := 1;  GOTO L5;
L3: IF !(TOF0__Q && !__timer_prev_TOF0) GOTO L5;
    IF !(TOF0__ET < TOF0__PT) GOTO L4;  TOF0__ET := TOF0__ET + 1;
L4: TOF0__Q := TOF0__ET < TOF0__PT;
L5: __timer_prev_TOF0 := TOF0__IN;
\end{lstlisting}
\caption{GOTO code that ESBMC-PLC emits for the timer rung of the running example. Names are as
\texttt{-{}-goto-functions-only} prints them; labels L1 to L5 replace \gls{esbmc}'s 19, 21, 22, 23, and 24, and the converter's no-op guards
\texttt{IF !1} and leading \texttt{1~\&\&} are omitted.}
\label{lst:tof}
\Description{A listing of GOTO instructions for one timer rung: the two power-flow accumulators, then assignments to the timer's input, elapsed time, output, and previous-input shadow.}
\end{figure}

\subsection{Graphical networks}
\label{sec:graph}

A graphical network is a directed graph of contacts, blocks, and coils, with wires read from left to right. Power
reaches a coil when some path from the left rail closes~\cite[\S8.2.3]{iec61131-3-2025}, so the coil's power flow is
the disjunction, over rail-to-coil paths, of the conjunction of the contacts on each path. Enumerating paths is
exponential in a bridge network. The resolver instead runs a depth-first search back from each coil and gives every
contact $n$ it reaches an accumulator
\begin{equation}
\label{eq:pf}
\mathit{pf}_n \;=\; \Big(\bigvee_{q \in \mathit{pred}(n)} \mathit{pf}_q\Big) \wedge \mathit{cv}_n ,
\end{equation}
with the left rail true and a timer or counter contributing its $Q$ after its single step per scan. These are the
\texttt{pf} variables of \cref{lst:tof}; each coil, a sink of the graph, then becomes one rung whose accumulator is the
disjunction at its input. A block is stepped once, at the first coil that reaches it, and later coils read the $Q$ of that step.

The graph gives no order between coils. Beremiz, whose output MATIEC compiles, orders them by their position in the
drawing, and the resolver does the same; it refuses a layout in which coils lie in rows less than 10 coordinate units apart in a chain, where that
order is not well defined. No program of the suite has such a layout. It also refuses wire cycles and any block other than a timer or counter in power flow, which it could not
represent as a contact in \cref{eq:pf}.

\subsection{Timers in milliseconds}
\label{sec:duration}

By default, a timer counts scans: a running timer adds one to \textit{ET} per scan and a preset is a number of scans.
This is exact for a fixed task period and is what the theorem of \cref{sec:simulation} states. Controllers
measure milliseconds, and a scan lasts as long as the branches it takes. With \path{--ld-cycle-time} $T_{\min}$:$T_{\max}$,
each scan draws one duration $d \in [T_{\min}, T_{\max}]$ in milliseconds, shared by every timer in that scan, and a
running timer advances to
\begin{equation}
\label{eq:adv}
\mathit{ET}' = \min(\mathit{ET} + d,\; \mathit{PT}) \quad \text{when } \mathit{ET} < \mathit{PT}.
\end{equation}
Start, hold, and expiry conditions stay as in \cref{tab:translation}, and $T_{\min} = T_{\max} = 1$ gives back the
scan-count rules. The shared $d$ reflects how MATIEC's timer library and the runtimes we inspected read the clock;
\cref{sec:refinement} states that and three other conditions as preconditions of the result.

\subsection{Properties and verdicts}
\label{sec:monitors}

Properties are Boolean expressions over BOOL variables, written in YAML. \Cref{fig:props} gives the property file of
the running example, which states \cref{eq:p1}. Each entry has an identifier and a kind; a response or reachability
property must also carry a justification, the reason its bound or witness is expected.
\Cref{tab:monitors} gives the five kinds and the monitor each compiles to. Two definitions are easy to misread. Mutual
exclusion is pairwise, ``at most one'', not ``not all at once''. A response property bounds every \emph{consecutive}
run of scans in which the trigger holds without the response; in timing mode the bound is written \texttt{max\_ms} and the
monitor counter adds $d$ instead of one, capped at $D + 1$ so that it stays finite.

\begin{figure}[t]
\begin{lstlisting}
properties:
  - id: P1
    kind: invariant
    expression: "!stairs_light || stairs_pir_sensor || lights_buttons_state"
    description: "Light must not be on without PIR detection or button activation"
  - id: R1
    kind: response
    trigger: Start
    response: Out
    max_scans: 2
    justification: "direct coil; Out tracks Start in the same scan"
\end{lstlisting}
\caption{A property file. \texttt{P1} is \cref{eq:p1} for the running example; \texttt{R1}, from another program
of \gls{esbmc}'s regression tests, bounds a response by two scans. In timing mode \texttt{max\_ms} replaces
\texttt{max\_scans}.}
\label{fig:props}
\Description{A YAML file listing an invariant property and a response property.}
\end{figure}

\begin{table}[t]
\centering
\small
\caption{Property kinds and their monitors. $c$ is a counter local to the monitor, initially 0.}
\label{tab:monitors}
\begin{tabular}{@{}lll@{}}
\toprule
Kind & Monitor at each scan boundary & A counterexample is \\
\midrule
\texttt{invariant}($e$) & assert $e$ & a boundary where $e$ fails \\
\texttt{absence}($e$) & assert $\neg e$ & a boundary where $e$ holds \\
\texttt{mutual\_exclusion}($x_1, \dots, x_n$) & assert $\neg(x_i \wedge x_j)$ for all $i < j$ & two of them true together \\
\texttt{reachability}($e$) & if $e$ then assert false & a witness that $e$ is reached \\
\texttt{response}($t, r, D$) & $c := (t \wedge \neg r)\;?\;c + 1 : 0$; assert $c \le D$ & $t \wedge \neg r$ held for more than $D$ scans \\
\bottomrule
\end{tabular}
\end{table}

Each run returns one of four verdicts. SAFE means k-induction proved every monitor at every reachable scan boundary.
VIOLATION means \gls{bmc} or the base case found a finite input sequence that breaks a monitor, and \gls{esbmc} prints it as a
trace. UNKNOWN means the solver timed out or the induction did not close; a bounded run that finds no violation is also
UNKNOWN, because no bound covers a loop that never ends. REFUSED means the frontend rejected the program, and there
is no verdict at all.

A SAFE verdict says nothing when the condition a property guards against never occurs. ESBMC-PLC therefore derives,
from each property, the conditions that must be reachable for it to constrain anything (the antecedent of an
implication, each variable of an exclusion, the trigger of a response) and checks each as a \texttt{reachability}
property. It reports a SAFE verdict on a property with an unreachable condition as vacuous. Arithmetic overflow is
checked in a separate bounded run, because \gls{esbmc}'s \path{--overflow-check} disables the inductive step.

\section{Why the translation is correct}
\label{sec:correctness}

A verdict from ESBMC-PLC is a statement about a GOTO program. Three results relate that program to the diagram: a scan
semantics, a theorem that one loop iteration and one scan agree, and a refinement of MATIEC's timers. All three are
about the translation as \cref{tab:translation} defines it; whether the compiled tool implements that definition is
checked per program (\cref{sec:tv}).

\subsection{Scan semantics}
\label{sec:semantics}

IEC~61131-3 runs a program from a periodic task~\cite[\S6.8.2]{iec61131-3-2025} but does not define how a scan reads
inputs and writes outputs; we adopt the read, execute, and latch structure of MATIEC and OpenPLC. Appendix~\ref{app:iec2025}
maps the semantics, the translation, and the evaluation to the clauses of the 2025 edition they rest on, and says
where each follows the standard, departs from it, or refuses the construct. A state $\langle\sigma, \pi\rangle$ pairs the store
$\sigma$, which maps each declared variable, and each state pin and literal preset of a graphical block, to a value
of its type, with an auxiliary store $\pi = (\pi_e, \pi_f)$ of
what a scan remembers: $\pi_e(v)$ is the previous-scan value of each edge-sensed variable and $\pi_f(i)$ the previous
input of each timer or counter $i$. Given an input choice $c$, a scan replaces the inputs, runs the rungs, and latches
the edge-sensed variables:
\begin{equation}
\label{eq:scan}
\frac{\sigma_1 = \mathit{read}_c(\sigma) \qquad
      \langle R_j, \sigma_j, \pi_f^{j} \rangle \to \langle \sigma_{j+1}, \pi_f^{j+1} \rangle \;\; (1 \le j \le n) \qquad
      \pi_e' = \mathit{latch}(\sigma_{n+1})}
     {\langle P, \sigma, \pi \rangle \Longrightarrow_c \langle \sigma_{n+1}, (\pi_e', \pi_f^{n+1}) \rangle}
\end{equation}
For a fixed $c$ the relation is deterministic, so the only nondeterminism in a run is the sequence of input choices,
and a proof over all choices covers every environment.

Inside a rung the question is when contacts are read. The power flow into the $k$-th coil of a rung is the
conjunction of the contacts to its left, each evaluated in the store $\sigma_k$ current when that coil runs:
\begin{equation}
\label{eq:rung}
p_k \;=\; \textstyle\bigwedge \{\, \lfloor e_i \rfloor(\sigma_k, \pi) \;:\; i < k,\; e_i \text{ a contact} \,\}.
\end{equation}
Reading each contact at its own position instead gives a different answer when a rung writes a variable it reads. Two
rungs, \texttt{go -(S)- a} and \texttt{a -(R)- a -( )- b}, separate the readings: position-wise reading sets
\texttt{b}, \cref{eq:rung} does not. We compiled the graphical form of this program with Beremiz and MATIEC, and the
generated C reads \texttt{a} after the reset, as \cref{eq:rung} does. Each element steps by its rule
in \cref{tab:rules} (Appendix~\ref{app:rules}): each block reads its inputs from the store and records its input in $\pi_f$ at its own
step. Arithmetic wraps in two's complement in the operands' common type and is then reduced into the type of
\textit{OUT}, as the code computes it. A graphical network contributes one rung per sink,
in the drawing order of \cref{sec:graph}. The sink's power flow is $\mathit{PF}$ of its input: \cref{eq:pf} restricted to the
predecessors reachable from the left rail (\cref{eq:pf-lemma}), evaluated in the store at the start of the sink. A
timer or counter steps by its rule at the first sink that reaches it, in that same store and before the sink's coil,
and each input pin that power flow drives (\textit{IN}, \textit{CU}, \textit{CD}, $R$, or \textit{LD}) holds the
disjunction of $\mathit{PF}$ over the reachable predecessors wired to that pin.
The initial auxiliary store $\pi_0$ maps every shadow to false.

\subsection{One-scan equivalence}
\label{sec:simulation}

Each shadow of the semantics is a GOTO variable (\texttt{\_\_edge\_prev\_}$v$ for $\pi_e(v)$,
\texttt{\_\_timer\_prev\_}$i$ or \texttt{\_\_ctr\_prev\_}$i$ for $\pi_f(i)$), so a concrete store $s$ corresponds to a state, written
$s \sim \langle\sigma,\pi\rangle$, when they agree on every named symbol and on the generated state pins and presets of graphical blocks, up to
this renaming and the type map. The
theorem rests on eight hypotheses, H2 to H9 (Appendix~\ref{app:hyp}). \gls{esbmc} checks one on every run, no division by
zero (H2); the parser and type checker enforce three; three fix the configuration, among them one task and no user
function block; and H9 assumes that \gls{esbmc} gives the GOTO fragment its standard meaning. H1, no overflow, is not needed
for the theorem, because the semantics wraps as the code does; it relates that semantics to the standard, which makes
overflow an error, and \gls{esbmc} can check it only in a bounded run.

\begin{theorem}[One-scan equivalence]
\label{thm:scan}
Let $P$ be a program in the subset of \cref{fig:grammar} that the parser and type checker accept and that satisfies H3
to H9, let $s \sim \langle\sigma,\pi\rangle$, and let $c$ be an input choice, a value for each variable that
$\mathit{RI}$ assigns (\cref{sec:scan-encoding}), such that no division in this scan has a zero divisor (H2 for the
scan). Running $T_{\mathit{scan}}(P)$ from $s$ with $c$
yields $s'$ with $s' \sim \langle\sigma',\pi'\rangle$, where $\langle P,\sigma,\pi\rangle \Longrightarrow_c
\langle\sigma',\pi'\rangle$.
\end{theorem}

The proof inducts over the stages of \cref{eq:tscan}, with one lemma per element type (Appendix~\ref{app:proof}); it is
a paper proof, not mechanized. Its
contact and accumulator lemmas show that $E$ evaluates, in the store at the coil, to $p_k$ of \cref{eq:rung}; this is
where the reading rule matters, since the lemma does not hold for position-wise reading. Timer and counter lemmas
split on the arms of each block's code, with the guards on \textit{ET} and \textit{CV} keeping every increment in
range. For graphical networks a lemma on the resolver shows that, for an acyclic graph, the search terminates and each
$\mathit{pf}_n$ equals \cref{eq:pf} evaluated in the store at the start of its coil's emission.

By induction over scans, the scan-boundary states of the GOTO loop and of iterated \cref{eq:scan} coincide for every
input sequence up to the first division by zero. So a property proved on the GOTO program, with the H2 assertions, holds on every semantic trace, and
a counterexample that violates none earlier is a semantic trace (\cref{cor:traces,cor:verdicts}); UNKNOWN and REFUSED
carry no claim. The block and monitor invariants of research question RQ4 (\cref{sec:evaluation}) only assume a fact at the head of each scan and assert it
before the loop and at the end of each scan, and \cref{lem:inv} shows that they add no wrong SAFE verdict and no spurious violation: a SAFE verdict
proved with them is one for the program without them. In scan-count mode the corollaries therefore cover
\path{--ld-fb-invariants} too.

The semantics departs from MATIEC in five places, which Appendix~\ref{app:departures} lists. They do not affect the
theorem, which relates the GOTO program to the semantics, but they limit what a verdict says about the program MATIEC
runs; the one a reader meets first is that an edge contact compares with the value latched at the end of the previous
scan.

\subsection{Timers against MATIEC}
\label{sec:refinement}

Timing mode (\cref{sec:duration}) describes a real controller only under four preconditions:
\begin{enumerate}[label=\textbf{P\arabic*},leftmargin=*,nosep]
\item every timer call in a scan reads the same clock value;
\item every scan lasts a whole number of milliseconds in $[T_{\min}, T_{\max}]$, with $0 \le T_{\min}$, overruns
  included;
\item a TOF whose input rises while it runs resets, and a TP ignores its input during its pulse;
\item a preset is a constant, non-negative number of milliseconds that fits the type of \textit{PT} and \textit{ET}.
\end{enumerate}
P3 describes how MATIEC's TOF and TP behave, and the model behaves the same way, so the theorem below assumes only the
other three.
MATIEC's TON, TOF, and TP copy one global clock value at each call, which OpenPLC writes once per scan and Beremiz reads
once per cycle, so both runtimes satisfy P1. P2 depends on the runtime's clock. OpenPLC advances its clock by the
configured tick at every scan, whatever the scan really took, so it satisfies P2 with $T_{\min} = T_{\max}$ when the
tick is a whole number of milliseconds; Beremiz reads the system clock, whose differences have nanosecond resolution,
so P2 holds there over the measured range of scan times, but only up to rounding to the millisecond. The type checker enforces P4 by refusing a preset that is negative or that
an input, a shared variable, or the program can change.

\begin{theorem}[Cycle-boundary refinement]
\label{thm:refine}
Assume P1, P2, and P4, and run a MATIEC TON or TOF and the rule of \cref{eq:adv} from the initial state on the same
inputs, with $d_i$ the difference between the clock values of scans $i-1$ and $i$. After every scan they agree on $Q$,
\textit{ET}, and the previous input. A TP agrees on $Q$ and the previous input, and on \textit{ET} except between the first
scan with a low input at or after its pulse expires and the next trigger; there MATIEC resets \textit{ET} to
zero and the model keeps it at \textit{PT}.
\end{theorem}

\begin{figure}[t]
\centering
\fitwidth{\begin{tikzpicture}[x=1cm, y=1cm, font=\scriptsize\sffamily]
  \foreach \t/\i in {0.00/0, 0.80/1, 2.40/2, 3.20/3, 4.80/4, 6.00/5, 6.80/6, 8.40/7, 9.60/8, 10.40/9} {
    \draw[gray!50, dotted] (\t,0.45) -- (\t,-5.15);
    \node[below] at (\t,-5.15) {$T_{\i}$};
  }
  \draw[{Bar[width=4pt]}-{Bar[width=4pt]}, gray] (0.80,-5.75) -- node[below] {$d_2$} (2.40,-5.75);
  \node[font=\scriptsize\sffamily\bfseries, anchor=east] at (-0.75,-0.90) {TON};
  \node[anchor=east] at (-0.1,0.10) {$\mathit{IN}$};
  \node[anchor=east] at (-0.1,-0.50) {$Q$};
  \node[anchor=east] at (-0.1,-1.80) {$\mathit{ET}$};
  \draw[thick] (0.00,0.00) -- (0.80,0.00) -- (0.80,0.30) -- (9.60,0.30) -- (9.60,0.00) -- (11.20,0.00);
  \draw[thick] (0.00,-0.60) -- (4.80,-0.60) -- (4.80,-0.30) -- (9.60,-0.30) -- (9.60,-0.60) -- (11.20,-0.60);
  \draw[gray, dashed] (0,-1.00) -- (11.20,-1.00) node[right, text=black] {$\mathit{PT}$};
  \draw[thick] (0.00,-1.90) -- (2.40,-1.90) -- (2.40,-1.50) -- (3.20,-1.50) -- (3.20,-1.30) -- (4.80,-1.30) -- (4.80,-1.00) -- (9.60,-1.00) -- (9.60,-1.90) -- (11.20,-1.90);
  \node[below] at (0.80,-1.90) {$T_s$};
  \node[anchor=west] at (4.96,-1.45) {$\mathit{ET}_i = \min(\mathit{PT}, T_i - T_s)$};
  \node[font=\scriptsize\sffamily\bfseries, anchor=east] at (-0.75,-4.00) {TP};
  \node[anchor=east] at (-0.1,-3.00) {$\mathit{IN}$};
  \node[anchor=east] at (-0.1,-3.60) {$Q$};
  \node[anchor=east] at (-0.1,-4.90) {$\mathit{ET}$};
  \draw[thick] (0.00,-3.10) -- (0.80,-3.10) -- (0.80,-2.80) -- (6.80,-2.80) -- (6.80,-3.10) -- (9.60,-3.10) -- (9.60,-2.80) -- (11.20,-2.80);
  \draw[thick] (0.00,-3.70) -- (0.80,-3.70) -- (0.80,-3.40) -- (4.80,-3.40) -- (4.80,-3.70) -- (9.60,-3.70) -- (9.60,-3.40) -- (11.20,-3.40);
  \draw[gray, dashed] (0,-4.10) -- (11.20,-4.10) node[right, text=black] {$\mathit{PT}$};
  \fill[gray!20] (6.80,-5.12) rectangle (9.60,-3.98);
  \draw[thick] (0.00,-5.00) -- (2.40,-5.00) -- (2.40,-4.55) -- (3.20,-4.55) -- (3.20,-4.33) -- (4.80,-4.33) -- (4.80,-4.10) -- (9.60,-4.10) -- (9.60,-5.00) -- (10.40,-5.00) -- (10.40,-4.78) -- (11.20,-4.78);
  \draw[thick, dashed] (6.80,-5.00) -- (9.60,-5.00);
  \draw[thick, dashed] (6.80,-4.10) -- (6.80,-5.00);
  \node[anchor=south] at (7.60,-4.10) {model};
  \node[anchor=south] at (8.20,-5.00) {MATIEC};
\end{tikzpicture}}
\caption{\Cref{thm:refine} on two timers over scans of uneven length, each signal drawn at its value at the end of a
scan. Top: a TON with $\mathit{PT} = 45$~ms started at $T_s = T_1$; MATIEC and the rule of \cref{eq:adv} give the same
\textit{ET} on every scan, because the steps $d_i$ telescope to $T_i - T_s$. Bottom: a TP with $\mathit{PT} = 40$~ms
whose pulse expires at $T_4$. From $T_6$, the first scan after expiry with \textit{IN} low, MATIEC resets \textit{ET}
to 0 (dashed) and the model holds \textit{PT} until the next trigger at $T_8$; the shaded scans are the only ones on
which the two differ.}
\label{fig:refine}
\Description{Two timing diagrams over ten scan boundaries of uneven spacing. The top one shows a TON whose elapsed time rises in steps to the preset and drops when its input falls, identical for MATIEC and the model. The bottom one shows a TP whose elapsed time reaches the preset; after the input falls, MATIEC drops it to zero while the model holds it at the preset until the next rising input, and that interval is shaded.}
\end{figure}

The proof inducts over scans (Appendix~\ref{app:proof} gives the case analysis). For a TON whose input rises on the scan with clock value
$T_s$, MATIEC computes $\mathit{ET} = T_i - T_s$ until the preset elapses, and the model's $\min(\mathit{PT}, \mathit{ET}_{i-1} +
d_i)$ telescopes to $\min(\mathit{PT}, T_i - T_s)$ (\cref{fig:refine}, top); P1 makes the telescoping hold for every timer of the scan at once,
and P4 keeps \textit{PT} constant across it. TOF follows the same argument from the scan on which its input falls, and
TP from its trigger (\cref{fig:refine}, bottom).

The theorem is about one timer against \cref{eq:adv}, and it does not by itself carry a verdict to the program MATIEC
runs. Theorem~\ref{thm:scan} is stated for scan-count mode (H5), the timing-mode timer code and the millisecond
monitor have no lemma of their own, and the only link between MATIEC's program and the semantics is the per-program
check of \cref{sec:tv}, which covers one scan. The block and monitor invariants add nothing to this gap: by
\cref{lem:inv} a deadline proved with them is a deadline of the timing-mode program without them. A property over a BOOL variable that the program computes from a TP's
\textit{ET}, or that depends on another
departure of Appendix~\ref{app:departures}, can even hold in the model and fail on MATIEC. We therefore present the
refinement, the translation validation, and the runs below as evidence that timing-mode verdicts carry over to MATIEC,
not as a proof. The converse fails by design: a range of durations can refute what a fixed period proves.

P1 and P2 are necessary. Driving MATIEC's compiled timers with a clock read per call raised a proved bound
of 70~ms to 90~ms, and one 40~ms scan in a 10~ms task raised another from 50 to 80~ms
(Appendix~\ref{app:timing}). With the preconditions kept, MATIEC's timers and an independent implementation of
\cref{eq:adv} agreed on $Q$ in all 318,240 random scans we ran.

\subsection{Per-program translation validation}
\label{sec:tv}

Theorem~\ref{thm:scan} trusts that the implementation computes \cref{tab:translation}. Translation
validation~\cite{Pnueli1998translation,Necula2000tv} removes that trust one program at a time. Its reference is the C
that MATIEC generates from the same PLCopen file after Beremiz renders it as Structured Text; \gls{esbmc} already reads that C.
Option \path{--ld-tv-check} runs one scan of the \gls{ld} side and one call of the MATIEC program body as \gls{ssa}
equations in one solver context, ties them through a correlation file, and asks whether some compared variable can
differ. The query gives both sides the same inputs and the same arbitrary value for every state variable that the
correlation file lists, starts the rest from their initial values, and compares the listed variables after the scan.
For the suite, a script writes the correlation file from MATIEC's record of the program's variables, so every declared
variable is tied as an input, tied as a state variable, or compared. Only MATIEC's own temporaries stay out.
UNSAT means that the two sides agree after one scan for every input and every pre-state of the listed variables; SAT
returns the input and state on which they differ.

An UNSAT answer is only as good as the query, so every query runs twice. The second run flips one compared Boolean in
the reference after its scan, which a correct harness must report as SAT. Each query starts timer, counter, and edge shadows at their initial values,
so it does not check behavior that needs a previous scan. For timers, \cref{thm:refine} and the oracle runs cover that
behavior; for counters and edge contacts only the agreement of the two routes on the suite's labels does
(\cref{sec:eval-setup}). \Cref{sec:conclusion} discusses what a shared reference cannot catch.

\section{Evaluation}
\label{sec:evaluation}

We evaluate ESBMC-PLC on two corpora of different origin, plus generated timer programs and two parametric families. The
benchmark suite carries properties and expected verdicts, and 217 of its 241 \gls{ld} variants are ours; the 413 programs
mined from GitHub carry no labels, and none of them is ours. We use the suite to test verdicts and the GitHub corpus to test the
translation, and we ask six questions.
\begin{enumerate}[label=\textbf{RQ\arabic*},leftmargin=*,nosep]
\item Does the tool return the expected verdict, and what does it do with programs outside the subset?
\item Does the emitted GOTO program agree with MATIEC's C, program by program?
\item What does the frontend add over verifying MATIEC's C of the same programs directly, and does an independent \gls{plc}
  verifier agree with its verdicts?
\item Can it prove deadlines in milliseconds, and at what cost?
\item How many SAFE verdicts are vacuous, how often does input classification fall back on its heuristic, and what does
  a run cost?
\item What happens when a property depends on more than one scan of history?
\end{enumerate}

\subsection{Setup}
\label{sec:eval-setup}

Every ESBMC-PLC number comes from one build of our fork: \gls{esbmc} 8.5.0 plus 38 commits that are not on upstream master
(commit \texttt{5efb80de72}). Among them are the options of RQ2 and RQ4, and two commits port the parser refusals that
upstream \gls{esbmc} added in esbmc/esbmc\iss{8858} (Appendix~\ref{app:gaps}). It is a Release build with Z3 4.13.3, compiled by
Apple clang 21.0.0 and configured with CMake 4.4.3. The runs read suite release v1.2.0 (commit \texttt{6d1e3c3}), MATIEC
commit \texttt{7949c0b}, and Beremiz commit \texttt{df4370c}, with Python 3.9.6 for the harness scripts, on an Apple M5
Pro with 15 cores and 24~GB of memory under macOS 26.7.1, between October 22 and 23, 2026. An earlier run of both suite routes, on October 18 with a
previous build of the fork and suite commit \texttt{d08e3f8}, enters only the last row of \cref{tab:rq1} and the
run-to-run comparisons of RQ3 and of the reliability threat (\cref{sec:threats}). Each run had \gls{esbmc}'s
wall-clock limit of 120~s and no memory limit, and the times we report are CPU times. PLCverif (RQ3, RQ6) is its command-line
build of July 17, 2026 (command-line interface 1.0.0.202607171441), run on OpenJDK 27.0.1 with NuSMV 2.7.1 built from source. Appendix~\ref{app:accounting} maps every count
to its variants and engine, and the artifact holds the command and record behind each number.

The corpus is the IEC~61131-3 benchmark suite~\cite{plc-benchmark-suite}. Its \gls{ld} route reaches 241 variants of 139 tasks,
each with a property file and an expected verdict fixed before these runs; we corrected one of those labels after an
earlier run of the tool, as described below. We report two provenance axes. By program, 217
variants come from the suite's authors and 24 are third-party: 18 from the PLC-LD dataset of ladder logic bombs~\cite{Iacobelli2024plcld,plcld-dataset}, which are the unnumbered
versions of 10 of its 15 programs, 8 as legitimate and malicious pairs and 2 as the legitimate program alone;
3 Beremiz examples; 2 from \emph{Lessons in Electric Circuits}~\cite{kuphaldt-lessons-iv}; and 1 from another public source. By label, 213 variants carry an expected
verdict that two routes, the \gls{ld} frontend and Beremiz to MATIEC to C, agree on, and 28 do not. One of those routes is the
tool under test, so agreement with these cross-checked labels is not independent evidence; \cref{sec:threats} returns
to this. The suite's runner picks k-induction for a task expected SAFE and \gls{bmc} to 20 scans for one expected to fail,
so a wrong VIOLATION label would never meet a prover. We instead ran k-induction for the 225 variants with a safety
property; the 16 variants that check a loop budget (Appendix~\ref{app:evaluation}) keep the runner's choice. Each SAFE
verdict also gets a run with \path{--overflow-check} to check H1, bounded at 20 scans because the flag turns off the
inductive step.

To test the frontend on programs nobody on our team wrote, we also searched GitHub on October 19, 2026, for PLCopen XML
files with a left power rail. Leaving out our own repositories gave 1,290 files from 173 repositories, and merging
copies that agree on their types and instances gave 1,018 distinct programs with an \gls{ld} body. We report two groups
apart: 413 programs from 138 repositories, mostly teaching, laboratory, and hobby projects for OpenPLC and Beremiz, and
605 from three security and clone-detection datasets, one of them a copy of PLC-LD. The other 32 repositories hold
only copies of programs credited to another (21) or no \gls{ld} body (11). Two of the 413 and 18 of the 605
are also in the suite. These programs carry no properties or labels, so they enter RQ2 only.

\subsection{RQ1: verdicts}
\label{sec:rq1}

\begin{table}[t]
\centering
\caption{Outcome on the 241 \gls{ld} variants against the labels of suite release v1.2.0, and, in the last row, against the
labels as they stood before one was corrected. ``Correct'' means the verdict equals the expected one and, for SAFE, that the
overflow run finds no overflow within 20 scans (H1). ``H1 violated'' is a SAFE verdict that rests on an overflow. Variants with a user
function block lie outside \cref{thm:scan}.}
\label{tab:rq1}
\small
\begin{tabular}{@{}lrrrrr@{}}
\toprule
& Correct & H1 violated & Refused & Unknown & Wrong \\
\midrule
Authored programs (217) & 207 & 0 & 10 & 0 & 0 \\
Third-party programs (24) & 10 & 7 & 7 & 0 & 0 \\
\midrule
Cross-checked labels (213) & 204 & 6 & 3 & 0 & 0 \\
Candidate labels (28) & 13 & 1 & 14 & 0 & 0 \\
\midrule
Without a user function block (220) & 208 & 0 & 12 & 0 & 0 \\
With a user function block (21) & 9 & 7 & 5 & 0 & 0 \\
\midrule
All (241) & 217 & 7 & 17 & 0 & 0 \\
All, labels before the correction (241) & 216 & 7 & 17 & 0 & 1 \\
\bottomrule
\end{tabular}
% single-build-2026-10-22/routes/uniform_ld (verdicts); route-runs-2026-10-18 records for the labels before the correction; single-build-2026-10-22/paper/suite_tables.txt for H1 and provenance
\end{table}

\Cref{tab:rq1} shows no wrong verdict and no UNKNOWN against the current labels. The 217 correct verdicts split into 127 SAFE, each proved by
k-induction at $k = 2$, and 90 violations: 83 found in the base case of k-induction and 7 loop-budget overruns found by
\gls{bmc} to 20 scans. Only 13 of the SAFE proofs involve timer or counter
state: 8 timer and 5 counter programs. The other 114 hold Boolean state alone, among them the 2 edge-block programs
of \cref{tab:plcverif}. All 17 refusals are programs outside the subset, and each diagnostic names
the construct. Twelve are variants of tasks the suite declares as \gls{fbd} or \gls{sfc}, and one is an \gls{fbd} or
\gls{sfc} body in a task declared \gls{ld}. Three put a comparison block in power flow, and one wires an arithmetic
output to an output variable. Seven SAFE verdicts, all on third-party programs whose user function block computes \texttt{IN1 - 5} on an
unconstrained INT, rest on an overflow that the second run finds; we do not count them as correct. Against the runner's own configuration, uniform k-induction changes one verdict. A forward and reverse
motor interlock from a textbook was labeled as failing and was UNKNOWN under 20-scan \gls{bmc}. Under k-induction it is SAFE
at $k = 2$, and \gls{esbmc} on MATIEC's C of the same program agrees (RQ3). On a \gls{plc} it is safe for a reason the relay
drawing does not show. The second rung reads the first coil after the first rung has written it, so the drawing
order, not the wiring, decides which motor direction wins when both are requested. The label came from a
counterexample that a frontend reading each coil's previous-scan value had produced, and release v1.2.0 of the suite
(October 18, 2026) corrects it. We made that correction after an earlier run of the tool had proved the program SAFE. The last row of \cref{tab:rq1} grades the same verdicts against
the labels as they stood, and there one verdict is wrong. We count it as correct because a route that does not use our
frontend, \gls{esbmc} on MATIEC's C, also proves the program SAFE. Without user function blocks, the tool
agrees with every label; nine in ten labeled variants come from the suite's authors, so this says nothing yet about industrial code.
The programs mined from GitHub (\cref{sec:rq2}) test the translation on third-party code, but not verdicts.

\subsection{RQ2: translation validation}
\label{sec:rq2}

\begin{table}[t]
\centering
\small
\caption{Translation validation over the 229 variants of tasks that the suite declares as \gls{ld}; the other 12 are \gls{fbd}
or \gls{sfc} bodies, which the frontend refuses.}
\label{tab:tv}
\begin{tabular}{@{}lr@{}}
\toprule
Outcome & Variants \\
\midrule
Equivalent after one scan (UNSAT) & 210 \\
Different (SAT), all through \texttt{EN}/\texttt{ENO} of a user function block & 13 \\
Refused by the frontend & 5 \\
Reference C did not build & 1 \\
\midrule
Mutation control reported SAT & 223 of 223 \\
\bottomrule
\end{tabular}
% single-build-2026-10-22/paper/logs/tv.log, logs/tv_control.log
\end{table}

\Cref{tab:tv} gives the outcome. Every difference has one cause: when a user function block's \texttt{EN} is false,
MATIEC skips the block and its outputs keep their values, while the frontend ignores \texttt{EN}. All 13 programs come
from the PLC-LD dataset, and all use user function blocks, which \cref{thm:scan} does not cover. The mutation control, a
flipped output in the reference, returned SAT on every query it ran, so no UNSAT answer comes from a harness that cannot
see a difference. During development the same check, with the timing model of \cref{sec:refinement} and probe programs
compiled by MATIEC, caught defects in the frontend's handling of INT width, REAL precision, and the
enable pins of user function blocks;
Appendix~\ref{app:defects} lists them with their fixes.

\begin{table}[t]
\centering
\small
\caption{Translation validation on the \gls{ld} programs mined from GitHub.}
\label{tab:tv-mined}
\begin{tabular}{@{}lrr@{}}
\toprule
Outcome & Projects & Datasets \\
\midrule
Distinct \gls{ld} programs & 413 & 605 \\
Accepted by the frontend & 115 & 39 \\
\quad Equivalent after one scan (UNSAT) & 104 & 2 \\
\quad Different (SAT) & 0 & 37 \\
\quad Reference or harness failed & 11 & 0 \\
\midrule
Mutation control reported SAT & 104 of 104 & 2 of 2 \\
\bottomrule
\end{tabular}
% single-build-2026-10-22/thirdparty/results.json; group split as in ws_thirdparty/REPORT.md
\end{table}

\paragraph{Programs mined from GitHub}
\Cref{tab:tv-mined} gives the outcome. The frontend accepts 115 of the 413 project programs, from 48 repositories. On
104 of them the query returns UNSAT and the control SAT, and no project program differs from MATIEC. The 37
differences in the datasets all lie in PLC-LD copies with user function blocks, the cause above. On 11 project
programs the harness failed, not the tool (Appendix~\ref{app:harness}). The 104 equivalent programs are small, with a
median of 6 contacts, coils, and blocks and at most 77; 27 use a timer or counter.

The subset accepts 28\% of the 413 project programs. Of the 298 it refuses, 2 contain an \gls{ld} jump or label, and 58 put an operator block such as ADD or
MOVE on a data wire; each refusal names its construct, so the list of causes in Appendix~\ref{app:refusals} is also
where the subset should grow.

\subsection{RQ3: the frontend against MATIEC's C and against PLCverif}
\label{sec:rq-c}

\gls{esbmc} reads the C that MATIEC generates, so a user could skip the \gls{ld} frontend and verify that C. The suite already has
this route (PLCopen XML to Beremiz Structured Text to MATIEC C, with a generated harness); we gave it the same scan model
as the \gls{ld} route, an unbounded scan loop with nondeterministic inputs and a clock that advances one millisecond per scan,
and the same engine and limit, k-induction with 120~s. We converted plain-text \gls{ld} variants to PLCopen XML first, as
for the \gls{ld} route, and the 16 variants of loop-budget tasks have no counterpart in the C harness.

\begin{table}[t]
\centering
\small
\caption{\gls{esbmc} on MATIEC's C, by the verdict of the \gls{ld} route, with k-induction and a 120~s limit on both routes, over
the 225 variants with a safety property. The 128 SAFE verdicts are the 134 of \cref{tab:rq1} without the 6 on
loop-budget variants; one of the 128 rests on an overflow (H1), and the C route times out on it.}
\label{tab:viac}
\begin{tabular}{@{}lrrrr@{}}
\toprule
& \multicolumn{4}{c}{MATIEC C route} \\
\cmidrule(l){2-5}
\gls{ld} route & SAFE & VIOLATION & Timeout or UNKNOWN & No C \\
\midrule
SAFE (128) & 2 & 0 & 125 & 1 \\
VIOLATION (83) & 0 & 82 & 1 & 0 \\
REFUSED (14) & 0 & 2 & 11 & 1 \\
\bottomrule
\end{tabular}
% single-build-2026-10-22/routes/viac/viac.jsonl against routes/uniform_ld
\end{table}

The two routes never disagree (\cref{tab:viac}), and they find the same violations: 82 of 83, with a median of
1.5~s per violation on the C route against 0.032~s on the \gls{ld} route. They part on proofs. Of the 128 SAFE verdicts of the \gls{ld}
route at $k = 2$, 127 are correct (\cref{sec:rq1}). The C route proves 2 of the 128 and ends in a timeout or UNKNOWN on
125; one variant has no C. We merge the two outcomes because neither is a proof and the split between them is not stable:
between two runs of the C route, on two builds, 16 of the 125 moved from one to the other. The C route does reach the \gls{fbd} and \gls{sfc} programs that the
frontend refuses, and finds 2 violations among them, but it proves none.

\paragraph{Tuning the C route}
A longer limit does not help. At 900~s, with \gls{esbmc} killed 30~s later if its own limit had not stopped it, the C route
proves none of the 125: 95 end UNKNOWN and 30 time out, and 73 of the timeouts at 120~s become UNKNOWN, an inductive step
that fails rather than one that runs out of time. The cause lies in MATIEC's variable records. \texttt{\_\_SET\_VAR} skips
the write when a variable carries MATIEC's force flag, and the inductive step may start from a state with that flag
set, so a forced output keeps an arbitrary value. Nothing in a program sets the flag. We therefore added, for every
variable, the fact that it is not forced, assumed at the head of the scan loop and asserted at its end, and for counter
and edge blocks the fact that the block is enabled. With these facts \gls{esbmc} proves 117 of the 125 on MATIEC's C, all at
$k = 2$: 112 with the force-flag facts alone and 5 counter programs with the enable facts as well. Of the 8 it still
cannot prove, 5 are among the 8 programs with a TON or TOF, which time out with or without facts on their timer
records, and 3 are Boolean programs. What the frontend adds, then, is proof without facts about a compiler's data
structures, and proof of the TON and TOF programs that the C route does not reach even with them. Its refusals and
warnings also speak in terms of the diagram.

\paragraph{An independent \texorpdfstring{\acrshort{plc}}{PLC} verifier}
To compare verdicts with a tool that shares neither the frontend nor \gls{esbmc}, we ran PLCverif~\cite{Darvas2015plcverif}
with NuSMV on the \gls{ld} variants within \cref{thm:scan}'s scope. PLCverif reads Siemens' \gls{scl}, so a script turns the
Structured Text that Beremiz generates for each variant, the same text the C route starts from, into an \gls{scl} function block
whose inputs are the variables the C harness drives each scan. Each property becomes an assertion at the end of the scan,
encoded as the C route encodes it, and each required witness (a \texttt{reachable} property) becomes a run of its own
whose assertion fails once the state is reached. PLCverif's S7-300 library supplies TON, TOF, and TP; for CTU, CTD,
R\_TRIG, and F\_TRIG we transcribed the IEC~61131-3 definitions into \gls{scl} and checked each block against one property it
must satisfy and one it must violate. \Cref{tab:plcverif} gives the outcome on the 207 variants this covers: the 224 the
frontend accepts, without the 16 that call a user function block and one for which Beremiz generates no Structured Text.
PLCverif agrees with every label and with every verdict of ESBMC-PLC, timer and counter programs included, and each run
takes at most 2~s. These labels therefore do not rest only on routes that share our frontend or \gls{esbmc}. Two C verifiers gave no usable comparison: 2LS called 10 harnesses with a known violation safe, and
CPAchecker could not parse MATIEC's variadic library functions (Appendix~\ref{app:second-verifier}).

\begin{table}[t]
\centering
\small
\caption{PLCverif with NuSMV against ESBMC-PLC on 207 of the 208 \gls{ld} variants within \cref{thm:scan}'s scope; Beremiz generates no Structured Text for the other.}
\label{tab:plcverif}
\begin{tabular}{@{}lrrr@{}}
\toprule
Variants & SAFE proved & VIOLATION found & Disagreements \\
\midrule
Boolean, clean and bomb pairs (160) & 80 of 80 & 80 of 80 & 0 \\
Boolean, other variants (30) & 29 of 29 & 1 of 1 & 0 \\
With TON, TOF, or TP (10) & 8 of 8 & 2 of 2 & 0 \\
With a counter or edge block (7) & 7 of 7 & none & 0 \\
\midrule
All (207) & 124 of 124 & 83 of 83 & 0 \\
\bottomrule
\end{tabular}
% results/plcverif-{bool,timers,counters,rest}-2026-10-23/results.tsv and witness.tsv
\end{table}

\subsection{RQ4: deadlines in milliseconds}
\label{sec:deadlines}

We generated 26 timer programs (TON, TOF, and TP with presets of 50 and 100~ms over four scan ranges, and a two-timer
cascade over two scan ranges) and computed each worst-case response by exhaustive search over an independent model of \cref{eq:adv}; for one
timer it is $T_{\max}$ plus the largest sum of durations below \textit{PT}. In all 26 cases a bounded run of ESBMC-PLC
over the worst-case horizon confirmed the predicted bound and refuted one millisecond less. On the running example,
with presets scaled down from 20~s, it confirmed both sides for six parameter sets: \textit{PT} after the last motion
with a fixed 20~ms scan (\textit{PT} of 200, 500, 1,000, and 2,000~ms), and $T_{\max} + \mathit{PT} - 1$ with scans of 15 to
25~ms (\textit{PT} of 100 and 150~ms).

\Gls{bmc} reaches these bounds only by unrolling every scan. With a fixed 10~ms scan it showed in 32~s (median of five runs,
interquartile range 31 to 33~s) that no run of up to 201 scans breaks a 2~s deadline; with scans of 10 to 30~ms it timed out after 300~s at 9 scans. k-induction needs
invariants instead. The block invariants of \path{--ld-fb-invariants}, such as $0 \le \mathit{ET} \le \mathit{PT}$ and
$Q \Rightarrow \mathit{ET} \ge \mathit{PT}$ for a TON with a constant preset, prove $\mathit{ET} \le \mathit{PT}$ at
$k = 2$ for presets up to 20~s, where every case is UNKNOWN without them. The tool asserts every fact before it assumes
it, so a wrong fact yields a violation, never a SAFE verdict (\cref{lem:inv}). With one more fact from a prototype option,
$c \le T_{\max} + \mathit{ET}$ while the response monitor counts, k-induction proved the exact deadline at $k = 2$ for
TON, TOF, and TP with scans of 10 to 30~ms and presets up to 20~s, each in under 0.2~s. That fact covers one timer
whose trigger and response are its own variables; a cascade, or the running example, still needs \gls{bmc}.

\subsection{RQ5: vacuity, input classification, and cost}
\label{sec:rq4}

The vacuity check of \cref{sec:monitors} covered 207 invariant, exclusion, and response properties and found 2 vacuous:
a fault latch that a bomb variant never reaches and a second pump that an alternation program never runs. Both are
facts about the programs. Over the 224 variants the frontend reads (the 229 of \cref{tab:tv} without 5 refusals), 636 of 1,133 declared
variables are inputs by their
interface section, 8 by the use rule, and none by address alone, so the heuristic decides 8 variables in the corpus. The
median run takes 0.031~s of CPU time (interquartile range 0.025 to 0.032~s) and the longest 0.36~s; the programs are
small, and RQ4 and RQ6 show where the cost grows.

\subsection{RQ6: properties that need more than one scan}
\label{sec:deep}

Every SAFE verdict of RQ1 closes at $k = 2$. The cause lies in \gls{esbmc}'s inductive step, not only in the suite. Since
esbmc/esbmc\iss{2328}, which fixed a wrong verdict of k-induction, the step checks the claim in every unrolled iteration and
assumes it only after checking it. The claim of the first iteration is then checked from an arbitrary state, so the
step fails at every $k$ whenever one scan from such a state can break the property. A property closes only if one scan
from any state preserves it, and \gls{esbmc} first runs the step at $k = 2$. Every suite property is of
that kind. To measure what happens beyond such properties, we generated two families of text-\gls{ld} programs, each with a safe
variant and a bomb, for a size $n$.

\begin{itemize}[leftmargin=*]
\item \emph{Delay chain.} Two redundant channels pass one input through $n$ latch stages, with the rungs from stage $n$
  down to stage 1, so the input reaches stage $n$ after $n$ scans; the property is that the outputs agree. It is
  k-inductive for $k = n$. The bomb drops one channel's middle stage while a trigger input is set.
\item \emph{Dual counter.} Two CTUs count the same pulses to the preset $n$; the property is that their outputs agree.
  It is not k-inductive for any $k$, since unequal counts can wait without pulses before one pulse exposes them; a proof
  needs the invariant that the counts are equal. The bomb gives one counter the preset $n + 1$.
\end{itemize}

\begin{table}[t]
\centering
\small
\caption{The two families with ESBMC-PLC (k-induction up to $k = 64$, 300~s) and with PLCverif and NuSMV. A
violation found at $k$ means a counterexample of $k$ scans.}
\label{tab:deep}
\begin{tabular}{@{}llll@{}}
\toprule
Family, $n$ & Engine & Safe variant & Bomb \\
\midrule
Delay chain, 2 to 16 & ESBMC-PLC & UNKNOWN & found at $k = n$ (at most 0.16~s) \\
 & PLCverif + NuSMV & proved (1~s) & found (1~s) \\
Dual counter, 2 to 24 & ESBMC-PLC & UNKNOWN & found at $k = 2n - 1$ (0.05 to 50~s) \\
 & ESBMC-PLC, \path{--ld-fb-invariants} & UNKNOWN & same \\
 & PLCverif + NuSMV & proved (16~s) & found (1~s) \\
\bottomrule
\end{tabular}
% results/deep-benchmarks-2026-10-23: esbmc_plc_dual_chain.jsonl, pinned_dual_counter.jsonl, fbinv_dual_counter.jsonl, plcverif*/results.txt
\end{table}

\Cref{tab:deep} gives the outcome. ESBMC-PLC finds every bomb, and the length of the counterexample grows with $n$, to 47
scans for the counter with $n = 24$; it proves no safe variant. The block invariants of RQ4 do not help on the counters,
because they constrain each counter alone, not the relation between two. PLCverif with NuSMV, which computes the
reachable states rather than inducting over scans, proves every safe variant. To check the cause, we built \gls{esbmc} with
the claims of earlier iterations assumed instead of checked, as before \iss{2328}: it proves every delay chain at $k = n$,
still none of the counters, and leaves all 241 suite verdicts unchanged, but it also brings back the wrong verdict that
\iss{2328} fixed, so we use it only as a diagnostic. Properties that depend on history therefore need invariants, as the
deadlines of RQ4 do, or an engine that computes reachable states.

\subsection{Threats to validity}
\label{sec:threats}

\paragraph{Internal validity}
The labels are the main threat. The 213 cross-checked labels agree across our frontend and the MATIEC route, so on their
own they are not independent of the tool we evaluate. PLCverif confirms the label of every one of the 207 variants it
covers (\cref{tab:plcverif}) without our frontend, MATIEC's C, or \gls{esbmc}; it still shares Beremiz's Structured Text
generator with the C route, so a defect there could mislead both. The other 19 cross-checked labels rest
on the two routes alone. One label changed after an earlier run of the tool, and against the earlier labels the tool returns one wrong
verdict (\cref{tab:rq1}). On the 16 loop-budget variants, which keep the runner's engine choice, every \gls{bmc} run found
the overrun its label predicts. The deadline proofs of RQ4 run in timing mode, which \cref{thm:scan} does not cover (H5); the
block and monitor invariants they add introduce no wrong SAFE verdict and no spurious violation (\cref{lem:inv}), but the monitor invariants are not yet
upstream. Translation validation compares one scan from initial shadows, so it cannot see a difference that only a
previous scan's timer or counter state would expose. The comparison with PLCverif rests on our conversion to \gls{scl} and on
the four IEC blocks we transcribed; each block passed a property it must satisfy and one it must violate.

\paragraph{Construct validity}
The properties of nine in ten variants are ours. Every SAFE proof closes at $k = 2$ because one scan from any state
preserves every suite property, and \gls{esbmc}'s inductive step proves nothing else (RQ6); the suite therefore does not
exercise properties that depend on history, and the two families of RQ6 measure that gap only for two shapes of program.
``Correct'' for a SAFE verdict includes H1, which we check only to 20 scans.
The vacuity check finds 2 of 207 properties vacuous and does not cover another 144.

\paragraph{External validity}
We wrote 217 of the 241 labeled variants. The 24 third-party variants come from four sources, and the largest, the PLC-LD
dataset, is about logic bombs rather than safety. The programs mined from GitHub are mostly teaching and hobby projects,
the accepted ones have a median of 6 elements, and the subset accepts 28\% of them. We had no programs exported from
commercial environments.

\paragraph{Reliability}
We ran both suite routes twice, on October 18 and 22, 2026, with two builds of our fork; the \gls{ld} route
returned the same verdict on all 241 variants in both runs. Each run measured each variant once, so the times of RQ5
are single measurements; at a median of 31~ms of CPU time, they bound the cost but support no comparison of times. We repeated the experiments of RQ4, whose runs reach hundreds of seconds, five times: every one of their
120 configurations (28, 12, 20, and 60 in the four experiments) returned the same verdict in all five, and the 201-scan \gls{bmc} run had an interquartile range of 30.8 to
33.4~s.
\gls{esbmc}'s own time limit did not always stop a run on MATIEC's C at 900~s, so a run that exceeded the limit counts as a
timeout.

\section{Related work}
\label{sec:related}

\paragraph{Formal semantics of \gls{plc} languages}
Formal treatments of IEC~61131-3 programs go back more than two decades~\cite{Frey2000formal}, with model checking of
Instruction List~\cite{Canet2000il} among the first. K-ST~\cite{Wang2023} gives Structured Text an executable semantics
in the K framework~\cite{Rosu2010k}, so the definition runs programs and can be compared with a runtime. K-\gls{esbmc}~\cite{dantas2026kesbmc}
extends it to \gls{ld} as K-LD, an executable semantics of contacts, coils, the scan cycle, timers, counters, and edge
blocks, and we used K-LD as a differential oracle while building the frontend (Appendices~\ref{app:timing} and~\ref{app:defects}). Ebnenasir's
report~\cite{Ebnenasir2023tr} formalizes a subset of ladder logic as Promela templates and does not cover TOF, TP,
counters, arithmetic, edge-detecting contacts, or graphical networks (Appendix~\ref{app:ebnenasir}). Our semantics has a
narrower purpose: we wrote it against one translator, so that a theorem can relate the two, and we state each place where it
departs from the standard, as a hypothesis or as a listed departure (Appendices~\ref{app:departures}
and~\ref{app:iec2025}).

\paragraph{Verifiers for \gls{plc} programs}
Ovatman et al.~\cite{Ovatman2016} survey model checking of \gls{plc} software and classify the work by model checker and by
how the program is converted into the checker's input language; for relay ladder logic, Moon~\cite{Moon1994} gave an
early modeling technique that feeds such programs to a model checker.
PLCverif~\cite{Darvas2015plcverif,Darvas2014forte} translates \gls{plc} programs into the input languages of several model
checkers and has been applied to programs of industrial size~\cite{FernandezAdiego2015tii};
Arcade.PLC~\cite{Biallas2012arcade} is a verification platform for \gls{plc} programs, and other flows target
nuXmv~\cite{Cavada2014}. Each embeds a source-to-model translation whose fidelity the tool assumes. ESBMC-PLC checks that assumption. It also
reads graphical PLCopen XML directly and proves properties for every number of scans with k-induction. The PLCverif paper~\cite{Darvas2015plcverif} documents Siemens \gls{scl}, Siemens' implementation of Structured Text, as its
input, with \gls{sfc} and \gls{stl} partly supported, and does not mention \gls{ld} or PLCopen XML. We therefore compared verdicts
through a translation of each program's Structured Text into \gls{scl}, with the same properties, and PLCverif with NuSMV agreed
on all 207 variants it covered (\cref{sec:rq-c}). NuSMV computes the reachable states, which also proves the properties
that depend on history in \cref{sec:deep}; ESBMC-PLC's k-induction does not. We did not compare with Arcade.PLC or a
nuXmv flow of our own.

\paragraph{Translation validation}
Checking each run of a translator instead of the translator itself is translation
validation~\cite{Pnueli1998translation}, applied to optimizing compilers by Necula~\cite{Necula2000tv}. Verified
compilation~\cite{Leroy2009compcert} proves the translator once, at the cost of a mechanized semantics for source and
target; \gls{esbmc}'s GOTO language has none, so we took the per-run route. Differential testing compares implementations on
generated inputs, as Csmith did for C compilers~\cite{Yang2011csmith}. One property of \gls{plc} programs makes the per-run
check cheap here: a scan body is loop-free, so one \gls{smt} query covers every input of one scan. Resolving a graphical
network into ordered assignments resembles building an intermediate representation from a dependence
graph~\cite{Cytron1991ssa}; what \gls{ld} adds is power flow as a value, a coil order taken from the drawing, and blocks that
step once per scan, and we claim nothing new for the depth-first construction itself.

\paragraph{Initial preprints}
Initial versions of ESBMC-PLC and its K-LD oracle appeared in five arXiv preprints. The first~\cite{dantas2026esbmcplc} read textual
PLCopen XML, encoded the scan loop, and checked YAML properties by \gls{bmc} and k-induction; the
second~\cite{dantas2026graphplc} added the depth-first resolver for graphical networks (\cref{sec:graph}); the
third~\cite{dantas2026esbmcplcplus} added timers, counters, and edge blocks to graphical \gls{ld} and a Structured Text
route through MATIEC. K-LD is the semantics of K-\gls{esbmc}~\cite{dantas2026kesbmc}, which used it as a differential oracle
on 13 programs and, for the combinational and latch fragment, machine-checked in kprove that its rules match the
standard. That result relates the semantics to the standard, not the translator to the semantics, so
\cref{thm:scan} stays a paper proof. A fifth preprint~\cite{dantas2026benchmark} describes an initial release of the
suite, with 83 variants of 50 programs. None of the five states a subset that the frontend refuses outside, proves the
translation against a scan semantics, or checks it program by program against MATIEC's C; those, the 241-variant and
413-program evaluations, and the PLCverif comparison are the contributions of this paper.

\section{Conclusion and limitations}
\label{sec:conclusion}

ESBMC-PLC verifies ladder diagrams by translation into an existing model checker and makes that translation
checkable. The subset is explicit, and the frontend refuses what lies outside it, apart from the gap of
Appendix~\ref{app:gaps}; one theorem relates the emitted loop to a
scan semantics; the frontend models timers in milliseconds under stated conditions on the runtime's clock; and each program's
translation can be compared with MATIEC's in one \gls{smt} query. On 241 benchmark variants it proves 127 safe where
verifying MATIEC's C of the same programs proves 2 and returns no wrong verdict against the current labels. It also agrees with
PLCverif on all 207 of the 208 variants within the theorem's scope that PLCverif can run (Beremiz produces no Structured Text
for the other), agrees with MATIEC on all 104 checkable programs of 413 from
GitHub that we did not write, and on the running example finds both the property an engineer would get wrong and the
toggle that never works.

Five limits apply. The theorem, the timer refinement, and translation validation all use MATIEC as their reference,
so a defect that MATIEC and the frontend share, or one in \gls{esbmc}'s symbolic execution, would pass all three.
Translation validation covers one scan, so timer and counter behavior across scans rests on the refinement proof and
oracle runs. The evaluation shows agreement with labeled programs that are mostly ours, not effectiveness on
industrial code (\cref{sec:threats}). \gls{esbmc}'s inductive step proves only properties that one scan preserves, so a
property that depends on history stays UNKNOWN unless the user adds invariants (\cref{sec:deep}). And k-induction proves
deadlines over a range of scan times only for single timers.

Two steps would remove the main limits: a validation mode over two scans, which closes the gap for timers and
counters, and a reference runtime that does not use MATIEC, such as a commercial controller. A mechanized proof of
\cref{thm:scan} lies beyond this paper, and so do programs and properties from industrial users.

\section*{Data Availability}
The artifact~\cite{esbmcplc-artifact} holds the Dockerfile that builds the pinned ESBMC fork and the suite release, the
scripts behind every number, a table that maps each number to its record, and the records of the runs reported here.
The benchmark suite is archived separately~\cite{plc-benchmark-suite}. For the programs mined from GitHub, many of
which carry no license, the artifact holds the repository, path, commit, and blob of each file and a script that
fetches them, not the files.

\section*{Acknowledgments}
We thank the Department of Computer Science at the University of Manchester and its Systems and Software Security (S3)
Research Group for their support and computing resources. The Engineering and Physical Sciences Research Council (EPSRC) grants EP/T026995/1, EP/V000497/1, and
EP/X037290/1 and the Soteria project, awarded by UK Research and Innovation under the Digital Security by Design
(DSbD) Programme, funded this work in part.

\section*{Use of Generative AI}
The authors used Claude (Anthropic), a generative AI assistant, to draft and edit parts of the text; to write and
review parts of the tool code, test harnesses, and evaluation scripts; to run experiments under the authors' direction
and to trace each reported number to its record in the artifact's number-to-record table~\cite{esbmcplc-artifact}; to check citations; and to review the proofs before the co-authors checked them. The authors
reviewed and tested all of this material and take responsibility for it.

\bibliographystyle{unsrtnat}
\bibliography{refs,refs_self}

\appendix
\section{Proofs}
\label{app:proof}

This appendix states the correspondence, the corollaries, and the graph lemma of \cref{sec:simulation} in full and
proves them, and gives the case analysis behind \cref{thm:refine}.

Each shadow of the semantics becomes a GOTO variable: \texttt{ld::\_\_edge\_prev\_$v$} for $\pi_e(v)$ and
\texttt{ld::\_\_timer\_prev\_$i$} or \texttt{ld::\_\_ctr\_prev\_$i$} for $\pi_f(i)$. A timer or counter is named by
its instance, or by its $Q$ variable when the file gives no instance name. The correspondence $\sim$ between a concrete
store $s$ and an abstract state is then an identity up to renaming:
\begin{equation}
\label{eq:sim}
\begin{aligned}
s \sim \langle\sigma,\pi\rangle \iff\; & s(\texttt{ld::}v) = \sigma(v) \\
\wedge\; & s(\texttt{ld::\_\_edge\_prev\_}v) = \pi_e(v) \\
\wedge\; & s(\texttt{ld::\_\_timer\_prev\_}i) = \pi_f(i),
\end{aligned}
\end{equation}
with the same clause for counter shadows and the type map BOOL to \texttt{bool}, INT to a 16-bit and DINT and TIME to
a 32-bit signed integer (H9), and REAL to \texttt{double}. A timer or counter of a graphical network keeps its
state in pins the resolver generates, $i$\texttt{\_\_ET} or $i$\texttt{\_\_CV} and $i$\texttt{\_\_Q},
and $\sim$ equates these with the block's state in $\sigma$. A literal preset is a generated constant,
$i$\texttt{\_\_PT} or $i$\texttt{\_\_PV}, that the block's code reads and no code writes, and $\sim$ equates it
with the preset in $\sigma$. The concrete store also holds generated accumulators, the \texttt{pf} variables and the input pin of a
graphical block, and monitor counters. $T_{\mathit{scan}}$ reads no accumulator before writing it, and $\mathit{MON}$,
which reads and writes the monitor counters, writes no named symbol and no shadow, so the proof tracks $s$ restricted
to the named symbols, the shadows, and the generated state pins and presets.

\begin{corollary}[Traces]
\label{cor:traces}
Under H3 to H9, for every initial choice and every input sequence, the scan-boundary states of the GOTO loop and of
iterated \cref{eq:scan} are related by $\sim$ through every scan before the first one that divides by zero.
\end{corollary}

\begin{corollary}[Verdicts]
\label{cor:verdicts}
Let $\Phi$ be invariant, absence, exclusion, or reachability properties over the store at scan boundaries, or response
properties taken over the product with their monitor counter, and let $\Omega$ be the assertions of H2. For a
reachability property, ``proved'' means that the target is unreachable. A semantic trace is the sequence of states after
scans $1, 2, \dots$, the boundaries at which $\mathit{MON}$ checks $\Phi$; the initial state is not one of them.
\textup{(i)} If $\Phi \cup \Omega$ holds on every trace of the GOTO program, every semantic trace satisfies $\Phi$.
\textup{(ii)} A counterexample to $\Phi$ that violates no assertion of $\Omega$ earlier in the trace is a semantic trace.
A SAFE verdict is case (i), taking \gls{esbmc}'s verdicts as sound, which nothing here certifies
(Appendix~\ref{app:hyp}). A VIOLATION of $\Phi$ whose trace violates no earlier assertion of $\Omega$ is case (ii); a
violation of $\Omega$ reports a division by zero and makes no claim about $\Phi$. UNKNOWN and REFUSED carry no claim.
\end{corollary}

The options \path{--ld-fb-invariants} and \path{--ld-monitor-invariants} turn a program $L$, the scan loop with its
monitors, into $L_{\mathcal{F}}$ for a set $\mathcal{F}$ of facts. Each fact $F$ is a Boolean expression without side
effects: a block fact over the block's \textit{ET} or \textit{CV}, \textit{PT} or \textit{PV}, $Q$, and shadow, and a
monitor fact over these and its monitor's counter. $L_{\mathcal{F}}$ asserts every fact once before the loop, assumes
every fact at the head of the loop body, and asserts it again at the end of the scan: a block fact after the latch, a
monitor fact after its monitor's step. Between an end-of-scan assertion and the next loop head lie only monitor steps, each
of which writes its own counter and nothing else, and property assertions, which write nothing; no block fact reads a
monitor counter, and a monitor fact reads only its own.

\begin{lemma}[Invariant instrumentation]
\label{lem:inv}
Fix an initial store and the nondeterministic choices of every scan, and let $\mathcal{A}$ be the assertions of
$\mathcal{F}$ in $L_{\mathcal{F}}$.
\textup{(i)} Until an assertion of $\mathcal{A}$ fails, $L_{\mathcal{F}}$ passes through the stores $L$ passes
through, and each of its assumptions holds where it is evaluated.
\textup{(ii)} If the verifier proves a set of assertions $\Psi \cup \mathcal{A}$ on $L_{\mathcal{F}}$, every trace of
$L$ satisfies $\Psi$, and a counterexample of $L_{\mathcal{F}}$ to $\Psi$ that violates no assertion of $\mathcal{A}$
earlier in the trace is a trace of $L$.
\end{lemma}

The lemma holds for any scan body, in scan-count mode and in timing mode, and does not depend on the facts being
true: a fact that fails ends the run with a violation of $\mathcal{A}$, never with a SAFE verdict. With $\Psi = \Phi
\cup \Omega$, it carries \cref{cor:verdicts} from $L$ to $L_{\mathcal{F}}$, so in scan-count mode a SAFE verdict proved
with \path{--ld-fb-invariants} holds on every semantic trace. Monitor facts exist only in timing mode, which
\cref{thm:scan} does not cover; there the lemma shows only that the facts add no wrong SAFE verdict and no spurious violation of a
property to the timing-mode program.

\begin{lemma}[Graph to rungs]
\label{lem:g1}
Assume H8, let the parser represent or refuse every wire, let the wiring graph be acyclic, and let every sink have a
predecessor reachable from the left rail. The resolver's
emission for a sink $x$ terminates, and after the rungs of every contact $n$ it reaches, the store holds
$\mathit{pf}_n = \mathit{PF}(n)$, where
\begin{equation}
\label{eq:pf-lemma}
\mathit{PF}(n) \;=\; \Big(\bigvee_{q \in \mathit{live}(n)} \mathit{PF}(q)\Big) \wedge \langle n \rangle_{\sigma_x},
\end{equation}
with $\mathit{PF}$ of the rail true, $\mathit{PF}$ of a timer or counter its Q pin after the block's single step,
and $\sigma_x$ the store at the start of $x$'s emission. Before that step, each input pin $p$ of the block $b$ that power flow drives
holds $\bigvee_{q \in \mathit{live}(b,p)} \mathit{PF}(q)$, where $\mathit{live}(b,p)$ is the set of predecessors wired
to $p$ that are reachable from the left rail.
\end{lemma}

Here $\mathit{live}(n)$ is the set of predecessors of $n$ that are reachable from the left rail, and $\langle n
\rangle_{\sigma}$ the test of $n$ in store $\sigma$. \Cref{eq:pf} ranges over all predecessors; the two agree because
an unreachable predecessor is either a contact, whose $\mathit{pf}$ is false, or a block, which the parser refuses as an
undriven sink.

The invariant throughout is $s_j \sim \langle \sigma_j, \pi_j \rangle$ after each stage of \cref{eq:tscan}. We write
$\lfloor v \rfloor$ for the Boolean projection of a variable: $v$ itself for BOOL and $v \ne 0$ otherwise.

\begin{lemma}[Projection]
If $s \sim \langle\sigma,\pi\rangle$ then $\llbracket \lfloor v \rfloor \rrbracket s = \lfloor \sigma(v) \rfloor$.
\end{lemma}
\begin{proof}
Immediate from the definition of the projection and the type map of \cref{eq:sim}.
\end{proof}

\begin{lemma}[Contact]
If $s \sim \langle\sigma,\pi\rangle$, the test $\mathit{cv}$ that $T_{\mathit{rung}}$ adds for a contact evaluates in $s$
to the contact's value in $\langle\sigma,\pi\rangle$.
\end{lemma}
\begin{proof}
A normally open static contact has $\mathit{cv} = \lfloor v \rfloor$, and a normally closed one $\neg\lfloor v
\rfloor$; both follow from the projection lemma. A rising contact has $\mathit{cv} = \lfloor v \rfloor \wedge
\neg\mathit{prev}_v$, and $s \sim \langle\sigma,\pi\rangle$ gives $\llbracket \mathit{prev}_v \rrbracket s =
\pi_e(v)$, so $\mathit{cv}$ holds exactly when $v$ is true now and was false at the last latch, which is the rising-edge
rule. The falling case is symmetric. The parser also accepts an edge contact with normally closed polarity, which is not in
\cref{fig:grammar}, so \cref{thm:scan} does not cover it; its $\mathit{cv}$ negates the edge test, where Beremiz renders
the same contact as a plain negation, a departure that Appendix~\ref{app:departures} lists.
\end{proof}

\begin{lemma}[Accumulator]
Let $E_k$ be the accumulator when the $k$-th coil of a rung emits. For any $s_k \sim \langle\sigma_k,\pi_k\rangle$,
$\llbracket E_k \rrbracket s_k = p_k(\sigma_k, \pi_k)$ of \cref{eq:rung}.
\end{lemma}
\begin{proof}
Only contacts extend $E$, so $E_k$ is the conjunction of the tests of the contacts before coil $k$. Apply the contact
lemma to each conjunct in $s_k$. The lemma needs no relation between $s_k$ and the store at the contact's position,
which is why \cref{eq:rung} evaluates contacts in the store at the coil.
\end{proof}

\begin{lemma}[Coil]
From $s \sim \langle\sigma,\pi\rangle$ with $p = \llbracket E \rrbracket s$, each coil's code produces a store related to
the state its rule in \cref{tab:rules} gives.
\end{lemma}
\begin{proof}
An output coil emits $v := \mathit{cast}_v(p)$, and the rule writes $p$ converted to the type of $v$; the two agree for
BOOL and for the 0/1 conversion of INT, DINT, TIME, and REAL. A set coil emits \texttt{if} $p$ \texttt{then} $v :=
\mathit{cast}_v(\mathit{tt})$, which
is the set rule and its skip case; the reset coil is symmetric. Only $v$ changes, and no shadow does.
\end{proof}

\begin{lemma}[Timer]
\label{lem:timer}
From $s \sim \langle\sigma,\pi\rangle$, with $\sigma$ the store at the block's step, the code of a TON, TOF, or TP
block computes its rule in \cref{tab:rules}.
\end{lemma}
\begin{proof}
By H3, \textit{IN}, \textit{PT}, \textit{ET}, and $Q$ are distinct variables, so a write to \textit{ET} or $Q$ does not
change the \textit{IN} or \textit{PT} read afterward, and $h = \mathit{IN} \wedge \mathit{prev}_i$ has one value
wherever the code reads it.
\emph{TON.} If $h$, \textit{ET} increases by one when $\mathit{ET} < \mathit{PT}$ and is unchanged otherwise, then
$Q' = (\mathit{ET}' \ge \mathit{PT})$; this is the running arm. Otherwise, $\mathit{ET}' = 0$ and $Q' = \mathit{ff}$,
the reset arm.
\emph{TOF.} If \textit{IN}, then $\mathit{ET}' = 0$ and $Q' = \mathit{tt}$. If \textit{IN} is false, $Q$ is true, and
$\pi_f(i)$ is false, \textit{ET} takes the guarded increment and $Q' = (\mathit{ET}' < \mathit{PT})$. In every other
case both are unchanged.
\emph{TP.} If $Q$, guarded increment and $Q' = (\mathit{ET}' < \mathit{PT})$. Else if \textit{IN} and not $\pi_f(i)$,
$\mathit{ET}' = 0$ and $Q' = \mathit{tt}$. Else unchanged.
In all three, $\mathit{prev}_i := \mathit{IN}$ gives $\pi_f'(i) = \lfloor\sigma(\mathit{IN})\rfloor$, the value at the
block's own step. Only the block's \textit{ET}, $Q$, and shadow change.
\end{proof}

\begin{lemma}[Counter]
From $s \sim \langle\sigma,\pi\rangle$, the code of a CTU or CTD block computes its rule in \cref{tab:rules}.
\end{lemma}
\begin{proof}
By H3 the block's writes to \textit{CV} and $Q$ do not change the \textit{CU}, \textit{CD}, $R$, \textit{LD}, or
\textit{PV} it reads.
\emph{CTU.} The count arm fires exactly when \textit{CU} holds, $\pi_f(i)$ does not, and $\mathit{CV} < \mathit{PV}$,
adding one; the reset arm then sets $\mathit{CV} := 0$ when $R$ holds, so a scan in which both fire ends with
$\mathit{CV} = 0$, the priority the rule states. Then $\pi_f'(i) = \lfloor\sigma(\mathit{CU})\rfloor$ and $Q' = (\mathit{CV}' \ge
\mathit{PV})$.
\emph{CTD.} The load arm sets $\mathit{CV} := \mathit{PV}$, exact because \textit{CV} is at least as wide as
\textit{PV} (H3); otherwise the count arm fires when \textit{CD} holds, $\pi_f(i)$ does not, and $\mathit{CV} > 0$,
subtracting one. Then $\pi_f'(i) = \lfloor\sigma(\mathit{CD})\rfloor$ and $Q' = (\mathit{CV}' \le 0)$. An unwired \textit{PV} reads 0 on both
sides.
\end{proof}

\begin{lemma}[Arithmetic]
If a divisor is nonzero (H2), $\mathit{OUT} := \mathit{cast}(\mathit{IN1} \;\mathit{op}\; \mathit{IN2})$ computes
its rule in \cref{tab:rules}: the operation wraps in the operands' common type, and the cast reduces the result into the type of
\textit{OUT}. MOVE is the case $\mathit{OUT} := \mathit{cast}(\mathit{IN1})$.
\end{lemma}
\begin{proof}
Under H9 the fragment's operators are two's-complement at the operand width, and division truncates toward zero, as the
rule states. The cast reduces modulo $2^{16}$ or $2^{32}$, which is the rule's reduction. No hypothesis on overflow is
needed, because the rule wraps where the code does; H1 only says that the wrap never happens.
\end{proof}

\begin{lemma}[Rung and latch]
Each rung maps a related state to the related state that the composition of its elements' rules in \cref{tab:rules}
gives, and the latch yields $\pi_e' =
\mathit{latch}(\sigma_{n+1})$.
\end{lemma}
\begin{proof}
Induction over the elements of the rung: contacts emit \texttt{skip} and leave both sides unchanged, and the coil, timer,
counter, and arithmetic lemmas carry the other steps, with the accumulator lemma supplying $p$ from the current store. Arithmetic blocks step only in textual rungs: the
resolver refuses one in power flow or wired to a variable or preset, and emits nothing for one whose output is
unconnected, which writes no variable.
The latch assignments read variables and write shadows only, so their order does not matter, and each sets
$\mathit{prev}_v$ to $\lfloor\sigma_{n+1}(v)\rfloor$.
\end{proof}

\begin{proof}[Proof of Theorem~\ref{thm:scan}]
$\mathit{RI}$ assigns the input values of $c$, giving $\sigma_1 = \mathit{read}_c(\sigma)$. The rung lemma applies to each
rung in turn, and the latch lemma closes the scan. For a sink of a graphical network, Lemma~\ref{lem:g1}
takes the place of the accumulator lemma: the drive rung's accumulator holds $\mathit{PF}$ of the sink's input in the
store at the start of the sink, which is the semantic power flow, and the coil lemma applies to it. A timer or counter
first reached at that sink steps after its input pin is set as Lemma~\ref{lem:g1} states, and before that sink's
drive rung; the timer and counter lemmas apply, because until the drive rung only generated variables are written, so
the block reads the store at the start of the sink, as the rule does. The \texttt{pf} variables and input pins are
written before they are read in each scan, so they need no counterpart in the semantics.
$\mathit{MON}$ preserves the relation, as the correspondence above notes. The increments in the timer and counter lemmas
need no separate assumption: each is guarded ($\mathit{ET} <
\mathit{PT}$, $\mathit{CV} < \mathit{PV}$, $\mathit{CV} > 0$), so no increment passes the preset and no decrement
passes 0, and both bounds fit the type by H3. Of the hypotheses, the proof uses H3, H8, H9, and H2 for the scan; H4 to H7 enter through
Corollary~\ref{cor:traces}.
\end{proof}

\begin{proof}[Proof of Corollary~\ref{cor:traces}]
Induction on the number of scans from $s_0 \sim \langle\sigma_0,\pi_0\rangle$, which holds by H7 and H4 for the store
and because the code starts every shadow at false, as $\pi_0$ does. Theorem~\ref{thm:scan} applies to each iteration
before the first division by zero, because under H5 and H6 the loop body is $T_{\mathit{scan}}(P)$ and nothing else.
\end{proof}

\begin{proof}[Proof of Corollary~\ref{cor:verdicts}]
A property is a predicate on the store after the latch, and a response monitor's counter is part of that state,
updated after the latch. Corollary~\ref{cor:traces} relates the two store sequences on every trace prefix that divides by zero nowhere.
\emph{(i)} If $\Phi \cup \Omega$ holds on every GOTO trace, no GOTO trace divides by zero, so
Corollary~\ref{cor:traces} covers every GOTO trace. Every semantic trace is one of them up to $\sim$, because
\texttt{nondet} ranges over each whole type (H9), and so it satisfies $\Phi$. \emph{(ii)} A counterexample with no earlier violation of $\Omega$ is a GOTO trace
to which Corollary~\ref{cor:traces} applies.
\end{proof}

\begin{proof}[Proof of Lemma~\ref{lem:inv}]
\emph{(i)} Assertions and assumptions write nothing, so where an assumption holds, $L_{\mathcal{F}}$ executes the
statements of $L$ in the same order on the same store. We show by induction over scans that every assumption holds.
The store at the head of the first scan is the store at which $L_{\mathcal{F}}$ asserted $\mathcal{F}$ before the loop.
The store at the head of scan $k + 1$ differs from the one at which scan $k$ asserted a fact only in the counters of
the monitors that stepped after that assertion, and the fact reads none of them. So if no assertion of $\mathcal{A}$ has failed, each fact holds where the loop head
assumes it, the assumptions block nothing, and the induction goes through.
\emph{(ii)} Let $\tau$ be a trace of $L$, and suppose some fact of $\mathcal{F}$ is false at a point where
$L_{\mathcal{F}}$ asserts it; take the first such point. Before it every assertion of $\mathcal{A}$ holds, so by (i)
$L_{\mathcal{F}}$ follows $\tau$ to that point and the assertion there fails, a trace of $L_{\mathcal{F}}$ that violates
$\mathcal{A}$, which the proof rules out. Every fact therefore holds wherever $L_{\mathcal{F}}$ asserts it on $\tau$, so
by (i) $\tau$ is also a trace of $L_{\mathcal{F}}$, and as the assertions of $\Psi$ sit at the same points in both
programs, $\tau$ satisfies $\Psi$. For the second claim, a counterexample that violates no assertion of $\mathcal{A}$
before its violation of $\Psi$ passes every assumption by (i) and so follows $L$ throughout.
\end{proof}

\begin{proof}[Proof of Lemma~\ref{lem:g1}]
\emph{Termination.} The emission is a depth-first search with an in-progress set and a visited set; each node is emitted
at most once per emission, and a node met while in progress is a wire cycle, which the parser refuses.
\emph{Value.} Induct over the order of emission, a post-order of the search and so a topological order of the reachable
subgraph. For a contact $n$, every $q \in \mathit{live}(n)$ is a rail or was emitted before $n$, and $\mathit{pf}_q$ is
written only by $q$'s own rungs, once per emission, so by induction it holds $\mathit{PF}(q)$. The rungs of $n$ clear
$\mathit{pf}_n$ and set it when some $\mathit{pf}_q$ holds together with $n$'s test; the test reads $n$'s variable and,
for an edge, its shadow, neither of which is $\mathit{pf}_n$, so all of $n$'s rungs read one store. Hence,
$\mathit{pf}_n = \mathit{PF}(n)$. A timer or counter, whose $Q$ is a generated pin in a graphical network, is the contact case without a test for each
input pin that power flow drives: the pin's rungs read only the $\mathit{pf}$ of the predecessors wired to that pin,
after which the block step assigns $Q$. The sink's drive rungs follow the same pattern without a test, then apply the coil.
Inside a sink's emission only generated variables are written before the final drive rung: a contact names a declared
variable and never a block's pin, and the parser refuses a block output wired to a variable (\cref{tab:excluded}).
So every contact reads the user variables of $\sigma_x$. Unfolding the recursion gives the disjunction over rail-to-sink paths of the conjunction of
their contacts, where a path through a timer or counter contributes, in place of the contacts before the block, the
block's $Q$ from its single step; that step may have happened during an earlier sink's emission. The parser enforces
all four preconditions of the lemma: H8, the representation of every wire (Appendix~\ref{app:order}), an acyclic
graph, and a reachable predecessor for every sink.
\end{proof}

\paragraph{Why the per-pin equation and H3 are needed}
Two small programs show that neither condition is a convenience of the proof. In the first, a CTU with $\mathit{PV} = 3$
takes \textit{CU} from a contact on \texttt{a} and $R$ from a contact on \texttt{b}. The resolver gives each pin its own
accumulator, as Lemma~\ref{lem:g1} states: \textit{CU} is the \texttt{pf} of the \texttt{a} contact, and the reset arm
tests the \texttt{pf} of the \texttt{b} contact. One input per block, the disjunction over all live predecessors, would
make \textit{CU} equal to $a \vee b$. The two readings part on the inputs $(a, b) = (0, 1)$ and then $(1, 0)$. Per pin,
the first scan resets and the second sees a rising \textit{CU}, so $\mathit{CV} = 1$; with one input, \textit{CU} is
high on both scans, the second sees no edge, and \textit{CV} stays 0.

The second program is a TON whose \textit{IN} and $Q$ name one variable, which the type checker refuses. The code
assigns $Q := \mathit{IN} \wedge \mathit{prev}_i \wedge \mathit{ET} \ge \mathit{PT}$ and then $\mathit{prev}_i :=
\mathit{IN}$, so with the two pins aliased the shadow records the $Q$ just written instead of the input. The shadow
starts false and so stays false, and the timer never runs: with \textit{IN} held high and $\mathit{PT} = 2$, the rule of
\cref{tab:rules} sets $Q$ on the third scan, while the aliased code keeps $\mathit{ET} = 0$.

\paragraph{Case analysis for \cref{thm:refine}}
We compare MATIEC's TON, TOF, and TP (\texttt{iec\_std\_FB.h}) with the rule of \cref{eq:adv}, scan by scan, under P1,
P2, and P4, writing $T_i$ for the clock value of scan $i$, $T_0$ for the clock value at the initial state, and $d_i = T_i - T_{i-1}$.
\emph{TON.} On the scan where \textit{IN} rises, MATIEC records $T_s$ and leaves \textit{ET} at the 0 of the preceding
idle call; the model also has $\mathit{ET} = 0$. While \textit{IN} stays high, MATIEC sets $\mathit{ET} = \min(T_i -
T_s, \mathit{PT})$ and the model $\min(\mathit{ET}_{i-1} + d_i, \mathit{PT})$, which telescopes to the same value;
$Q$ is $\mathit{ET} \ge \mathit{PT}$ on both sides. When \textit{IN} falls, both reset \textit{ET} and $Q$.
\emph{TOF.} While \textit{IN} is high both hold $Q$ true and $\mathit{ET} = 0$. On the falling scan MATIEC records
$T_s$ and both keep $Q$ true and $\mathit{ET} = 0$; afterward the telescoping argument of TON applies until
$\mathit{ET} = \mathit{PT}$, when both clear $Q$. A rise while running resets both (P3).
\emph{TP.} On a rising \textit{IN} with $Q$ false, both set $Q$ and start from $\mathit{ET} = 0$; during the pulse
both ignore \textit{IN} (P3) and the telescoping argument applies until $\mathit{ET} = \mathit{PT}$, when both clear
$Q$. After expiry MATIEC resets \textit{ET} to 0 on the first scan with \textit{IN} low, while the model keeps
\textit{PT} until the next trigger, where both start again from 0. $Q$ and the previous input agree throughout.

\section{Semantics details}
\label{app:semantics}

\subsection{Hypotheses}
\label{app:hyp}

\Cref{thm:scan} holds under H2 to H9. The verifier checks H2, and the type checker and parser enforce H3, H4, and H8.
H5 to H7 fix the configuration. H9 is an assumption about \gls{esbmc} itself. H1 is not a hypothesis of the theorem: it
relates the semantics, which wraps as the code does, to the standard, which makes overflow an
error~\cite[\S6.6.2.5.9]{iec61131-3-2025}.

\begin{enumerate}[label=\textbf{H\arabic*},leftmargin=*]
\item No \texttt{+}, \texttt{-}, \texttt{*}, or \texttt{/} overflows the type it is computed in
  (asserted under \path{--overflow-check}, in a run of its own: the flag turns off \gls{esbmc}'s inductive step, so the
  check is bounded).
\item No division in the scan has a zero divisor (asserted by default; off only with
  \path{--no-div-by-zero-check}).
\item A block's pins that it writes while still reading others are distinct variables, and \textit{ET} and
  \textit{CV} are at least as wide as \textit{PT} and \textit{PV} (type checker).
\item Every initial value lies in the range of its type (type checker).
\item The loop body is $T_{\mathit{scan}}(P)$ and nothing else: \path{--ld-fault-injection} and
  \path{--ld-cycle-time} are off, and \path{--ld-scan-watchdog}, which the suite's runner passes, instruments only
  \texttt{WHILE} loops in user function-block bodies and so adds nothing to a program in the subset; timing mode has
  its own statement in \cref{sec:refinement}. \path{--ld-fb-invariants} and \path{--ld-monitor-invariants} add only
  assumptions and assertions of facts, which add no wrong SAFE verdict and no spurious violation of a property
  (\cref{lem:inv}).
\item One task, one program, no user function block. The evaluated build refuses a user function block wired into
  power flow (Appendix~\ref{app:gaps}). The evaluation reports every program with a user function block separately
  (\cref{sec:rq1}).
\item The initial store is the declared-value store, with each generated state pin at its block's declared initial
  state and each generated preset at its literal, plus an initial choice for externally written variables under
  \path{--ld-closed-world}, and every shadow starts false.
\item No declared or referenced name is a shadow or a generated variable, and no two blocks share a shadow or a name
  (parser: every name beginning with \texttt{\_\_} is reserved, and a declared name equal to a generated one is
  refused).
\item \gls{esbmc} gives the fragment its standard meaning: two's-complement arithmetic at the operand width, division
  truncating toward zero, and \texttt{nondet} ranging over the whole type. DINT and TIME use \gls{esbmc}'s \texttt{int},
  so the run keeps the default 32-bit width.
\end{enumerate}

H1 and H2 are properties of the run, so the verifier checks them: H2 alongside the user's properties, H1 in a bounded
run of its own (\cref{sec:eval-setup}). The block and monitor invariants assume each fact at the loop head and assert it
before the loop and at the end of every scan, so they cut no trace unless an assertion has already failed;
\cref{lem:inv} proves this and carries the verdicts of \cref{cor:verdicts} to runs with them. Nothing in this
paper certifies the model checker: neither the theorem nor the translation validation of \cref{sec:tv} looks
inside \gls{esbmc}.

\subsection{Element rules}
\label{app:rules}

\Cref{tab:rules} gives the step of each element as a function of the store $\sigma$ and the block shadows $\pi_f$ at
the element's position. A rung is the left-to-right composition of its elements' steps, and a contact's step changes
nothing. The power flow $p$ of a coil is $p_k$ of \cref{eq:rung} in a textual rung and $\mathit{PF}$ of its input,
\cref{eq:pf-lemma}, in a graphical network. There a timer or counter steps at the first sink that reaches it, before
that sink's coil, and each input pin that power flow drives holds the disjunction of $\mathit{PF}$ over the live
predecessors wired to that pin. A pin the table does not change keeps its value. Each timer also sets $\pi_f'(i) =
\lfloor\mathit{IN}\rfloor$, a CTU $\lfloor\mathit{CU}\rfloor$, and a CTD $\lfloor\mathit{CD}\rfloor$.

\begin{table}[t]
\centering
\caption{The rule of each element. $\lfloor\cdot\rfloor$ is the Boolean projection, and $\mathit{inc}$ is
$\mathit{ET} + 1$ when $\mathit{ET} < \mathit{PT}$ and $\mathit{ET}$ otherwise. Arm names in parentheses are the ones
the proofs of Appendix~\ref{app:proof} use.}
\label{tab:rules}
\small
\begin{tabularx}{\linewidth}{@{}l>{\raggedright\arraybackslash}X@{}}
\toprule
Element & Step \\
\midrule
coil $v$ & $\sigma'(v) = \mathit{cast}_v(p)$, with 0 and 1 for INT, DINT, TIME, and REAL \\
set, reset coil $v$ & $\sigma'(v) = \mathit{cast}_v(\mathit{tt})$, or $\mathit{cast}_v(\mathit{ff})$ for reset, if $p$;
  unchanged otherwise \\
TON $i$ & with $h = \lfloor\mathit{IN}\rfloor \wedge \pi_f(i)$: if $h$ (running), $\mathit{ET}' = \mathit{inc}$ and
  $Q' = (\mathit{ET}' \ge \mathit{PT})$; otherwise (reset), $\mathit{ET}' = 0$ and $Q' = \mathit{ff}$ \\
TOF $i$ & if $\lfloor\mathit{IN}\rfloor$, $\mathit{ET}' = 0$ and $Q' = \mathit{tt}$; else if $Q \wedge \neg\pi_f(i)$
  (running), $\mathit{ET}' = \mathit{inc}$ and $Q' = (\mathit{ET}' < \mathit{PT})$; otherwise unchanged \\
TP $i$ & if $Q$ (pulse), $\mathit{ET}' = \mathit{inc}$ and $Q' = (\mathit{ET}' < \mathit{PT})$; else if
  $\lfloor\mathit{IN}\rfloor \wedge \neg\pi_f(i)$ (trigger), $\mathit{ET}' = 0$ and $Q' = \mathit{tt}$; otherwise
  unchanged \\
CTU $i$ & $\mathit{CV}_1 = \mathit{CV} + 1$ if $\lfloor\mathit{CU}\rfloor \wedge \neg\pi_f(i) \wedge \mathit{CV} <
  \mathit{PV}$ (count), else $\mathit{CV}$; $\mathit{CV}' = 0$ if $\lfloor R \rfloor$ (reset), else $\mathit{CV}_1$;
  $Q' = (\mathit{CV}' \ge \mathit{PV})$ \\
CTD $i$ & $\mathit{CV}' = \mathit{PV}$ if $\lfloor\mathit{LD}\rfloor$ (load); else $\mathit{CV} - 1$ if
  $\lfloor\mathit{CD}\rfloor \wedge \neg\pi_f(i) \wedge \mathit{CV} > 0$ (count); else $\mathit{CV}$; $Q' =
  (\mathit{CV}' \le 0)$ \\
ADD, SUB, MUL, DIV & $\mathit{OUT}' = \mathit{red}(\mathit{IN1} \;\mathit{op}_w\; \mathit{IN2})$, where
  $\mathit{op}_w$ wraps in two's complement at the operands' common width $w$, division truncates toward zero, and
  $\mathit{red}$ reduces modulo $2^{16}$ or $2^{32}$ into the type of \textit{OUT} \\
MOVE & $\mathit{OUT}' = \mathit{red}(\mathit{IN1})$ \\
\bottomrule
\end{tabularx}
\end{table}

\subsection{Constructs outside the subset}

\Cref{tab:excluded} gives the reason for each exclusion and how often it occurred among the 241 \gls{ld} variants of the
evaluation (\cref{sec:rq1}). The reasons are of two kinds. Some constructs have a meaning the scan semantics cannot
express without a new rule, such as preemption between tasks or the step sequencing of an \gls{sfc} chart; others could be
modeled and simply are not yet. Neither kind is translated approximately (\cref{sec:subset}).

\begin{table}[t]
\centering
\caption{Constructs outside the subset, why, and how many of the 241 \gls{ld} variants of the suite each refused.}
\label{tab:excluded}
\small\begin{tabular}{@{}>{\raggedright\arraybackslash}p{0.39\linewidth}>{\raggedright\arraybackslash}p{0.4\linewidth}r@{}}
\toprule
Construct & Reason & Refused \\
\midrule
\gls{fbd} and \gls{sfc} bodies & \gls{sfc} step sequencing is not modeled; \gls{fbd} needs its own network semantics & 13 \\
Comparison and other non-timer blocks in power flow & not modeled; dropping them was unsound (\cref{sec:graph}) & 3 \\
Block outputs wired to output variables, blocks wired to presets & data-path wiring not modeled & 1 \\
Several tasks, interrupt tasks & one scan loop cannot model preemption & 0 \\
Negated and transition-sensing coils & not modeled & 0 \\
A coil wired on to further elements & passing power through a coil is not modeled & 0 \\
\texttt{EN} or \texttt{ENO} wired on a standard block & conditional execution not modeled & 0 \\
Unsigned, 8- and 64-bit types, strings, arrays, structures & no kind in the type map & 0 \\
\bottomrule
\end{tabular}
\end{table}

Structured Text programs do not reach the \gls{ld} frontend at all, and 36 of the suite's 175 tasks have no \gls{ld} variant. We
leave Structured Text out of this paper because \gls{esbmc} reaches it only through MATIEC's C, a route that none of our theorems covers.

A separate component that no theorem here covers translates the Structured Text bodies of user function blocks.

\subsection{Where the semantics departs from MATIEC and the standard}
\label{app:departures}

We read the semantics against the current edition of the standard, IEC~61131-3:2025~\cite{iec61131-3-2025}, and
against MATIEC, which implements the 2013 edition~\cite{iec61131-3-2013}; for the clauses we use, Annex~B of the 2025 edition lists no
change that the subset exercises (Appendix~\ref{app:iec2025}). Writing $T$ from the code forced four changes
to the semantics we started with, all in the direction of the code: the coil reads its contacts at the coil
(\cref{eq:rung}); a block's previous input is recorded at the block's step; TOF and TP guard the \textit{ET} increment;
and CTD has a load arm. We checked each against MATIEC's generated C or its library source.

Five differences from MATIEC remain. None affects \cref{thm:scan}, which relates the GOTO program to the semantics, but
each limits what a verdict says about the program MATIEC runs:
\begin{itemize}
\item an edge contact compares with the value latched at the end of the previous scan, where MATIEC compares with the
  previous evaluation of the element (esbmc/esbmc\iss{8211}); the two differ only when the scan writes the sensed variable
  after the contact;
\item REAL is a double, where MATIEC uses a 32-bit float (esbmc/esbmc\iss{8212});
\item the \texttt{EN} and \texttt{ENO} pins of a user function block are ignored (esbmc/esbmc\iss{8210}), and the same pins
  on a standard block are refused;
\item a TP keeps \textit{ET} at \textit{PT} after its pulse until the next trigger, where MATIEC resets it to 0 once the
  pulse has ended with \textit{IN} low (\cref{sec:refinement});
\item a contact that is both negated and edge-sensing negates the edge test, where Beremiz renders it as a plain negated
  contact; the parser accepts it, although \cref{fig:grammar} does not list it.
\end{itemize}
Upstream \gls{esbmc} has since fixed the first two (esbmc/esbmc\iss{8907}, \iss{8910}) and refuses a wired \texttt{EN} or \texttt{ENO} on
a user function block (\iss{8859}); the evaluated build branches from upstream before these changes, and this appendix
describes it.
In the first four, MATIEC follows the standard, so they are departures from IEC~61131-3:2025 as well: an edge contact
senses a transition between evaluations of the element (Table~74), REAL is single precision (\S6.4.2), a block whose
\texttt{EN} is false is not executed (\S6.6.1.5), and a TP's \textit{ET} returns to 0 when \textit{IN} is low after the
pulse (Figure~17). MATIEC shares two differences from the standard. A result stored in a narrower INT wraps
without an assertion, where the standard makes it an error~\cite[\S6.6.2.5.9]{iec61131-3-2025}. And CTU stops
counting at \textit{PV} and CTD at 0, where the standard counts to $\mathit{PV}_{\max}$ and $\mathit{PV}_{\min}$, limits it leaves to the implementer
(Table~46); $Q$ is the same,
but \textit{CV} differs once a CTU passes its preset. The frontend refuses three further differences rather
than assuming them: a coil, MOVE, or input that writes a timer's or counter's state variable besides the block; a TOF, TP or
counter whose state variable starts nonzero, which MATIEC's instance ignores; and an action in a program, which the
model would run on every scan.

\subsection{Coil order and a dropped block}
\label{app:order}

Which sink runs first is not in the graph. Exports also list coils on the right power rail, and an earlier version of
our resolver followed that list; Beremiz, whose generated code MATIEC compiles, ignores it and orders coils by their position
in the drawing. The frontend now does the same, and refuses a layout in which rows lie too close for that order to
be well defined. Beremiz also runs a block or output variable with a positive \texttt{executionOrderId} before the
rest, while coils always follow their position; the pinned frontend refuses a positive identifier (Appendix~\ref{app:gaps}),
so the two orders agree on every program it accepts. Every identifier in the suite is 0. Order matters most when one variable has both a set and a reset coil, since the second one powered in
a scan wins; the frontend warns when both can fire in the same scan, which is what happens in the running example
(\cref{sec:example}).

The lemma needs the parser to represent or refuse every wire. One case escaped the original resolver, which dropped from a sink's
predecessors, without a diagnostic, any block that is neither a timer nor a counter when the sink also had a live contact. A program computing \texttt{y = a OR (n > 3)} verified \texttt{!y || a}, which is
false. The parser refuses such a block (esbmc/esbmc\iss{8435}), and Lemma~\ref{lem:g1} holds only with that refusal.

\subsection{Inputs the frontend refuses, and the gap that remains}
\label{app:gaps}

The frontend refuses five inputs that it would otherwise model outside the theorem. First, in a rung given in the textual
PLCopen form, a timer or counter would read its \textit{IN}, \textit{CU}, or \textit{CD} variable and ignore the
contacts to its left, and a coil to its right would receive the contacts' power flow rather than the block's $Q$; the
graphical form, which every program of the suite uses, drives the block from its power flow as \cref{tab:translation}
states. Second, two response properties with the same identifier would share one monitor counter. Third, a sink
downstream of a user function block in power flow would read a variable named after the block's $Q$. The fourth is the
\texttt{executionOrderId} of Appendix~\ref{app:order}. The fifth covers the jump, label, and return elements of a
graphical network (\S8.1.6 and Table~72 of the standard): the frontend would read them as unconnected elements and drop
them, and so verify a program that jumps over a rung as if every rung ran. The evaluated build refuses all five (commits
\texttt{1d7ad4332b} and \texttt{5efb80de72}), and upstream \gls{esbmc} refuses them from commit \anonhash{ddae49e1a5}
(esbmc/esbmc\iss{8858}). None occurs in the suite. Two project programs mined from GitHub contain a jump or label, and the
frontend refuses both (\cref{sec:rq2}).

One gap remains, in the evaluated build and upstream. The frontend reads an \gls{ld} body that holds textual rungs as
textual, so a jump, label, or return placed beside those rungs, outside any rung, is dropped without a refusal. Such a
body mixes the textual and graphical forms, which only our own textual extension produces, and neither the suite nor the
programs mined from GitHub contain one.

\section{Timing evidence}
\label{app:timing}

\subsection{Clock sampling in two runtimes}

We checked P1 in two runtimes' source. MATIEC's \texttt{TON}, \texttt{TOF}, and \texttt{TP} bodies copy the global
\texttt{\_\_CURRENT\_TIME} at each call. OpenPLC~v3 writes that global once per scan, after the program runs, by adding
the task interval to it; no clock is read at all, so every scan lasts exactly the nominal period. Beremiz reads
\texttt{CLOCK\_REALTIME} once at the start of each cycle. Both satisfy P1. They differ on P2, and the difference decides
which runtime the range describes: OpenPLC is the instance $T_{\min} = T_{\max}$ and nothing else, so it cannot
validate the range at all, while Beremiz's durations vary and have nanosecond resolution. A Beremiz run is covered when
its measured cycle times are whole milliseconds inside $[T_{\min}, T_{\max}]$, and nothing in that runtime enforces
either condition.

\subsection{Negative controls}

P1 and P2 are necessary. For each of P1 and P2 we proved a response bound with the tool, then
drove MATIEC's compiled timer code with a synthetic clock that breaks the condition, and measured the real response
time with the same monitor the tool uses. \Cref{tab:negctl} shows the result.

\begin{table}[t]
\centering
\small
\caption{Negative controls. The tool proves the bound; MATIEC's timer code, run with a clock that breaks one
precondition, exceeds it. With the precondition kept, the same driver gives the bound.}
\label{tab:negctl}
\begin{tabular}{@{}llrrr@{}}
\toprule
Broken & Program, scan (ms) & Proved & Kept & Broken \\
\midrule
P1, clock read per call & TON(30)$\to$TON(40), 10 & 70 & 70 & 90 \\
P2, one 40~ms scan & TON(50), 10 & 50 & 50 & 80 \\
P2, nanosecond clock & TON(50), 10--20 & 69 & 69.0 & 69.9 \\
\bottomrule
\end{tabular}
\end{table}

A per-call clock costs two scans in the cascade, because each timer starts late inside its scan and is read early later.
An overrun beyond $T_{\max}$ moves the bound by the overrun. A clock finer than a millisecond moves it by less than one
tick, so a deadline proved in the model needs a margin of one clock tick when read as a hard limit. P3 and P4 have no
control; P3 is a definition, and the type checker enforces P4, which refuses a preset that is negative or that
an input, a shared variable, or the program can change.

\subsection{MATIEC oracle and counters}

The proof reads the library source, so we also ran the library. We compiled MATIEC's TON, TOF, and TP for 26 generated programs
and drove each with random input sequences and random durations in its range, plus fixed patterns. Over 318,240
scans, MATIEC and an independent Python implementation of the model agree on $Q$ in every scan and on \textit{ET} in
every scan except 35,246, all of them the TP case the theorem names: the oracle counts every scan from the pulse's end
to the next trigger, not only the expiry scan. % paper-pin run: logs/l1_oracle_a.log
The clock in this run is synthetic. We ran no runtime with a wall clock, so the range has no oracle beyond the library
source and this run; K-LD and OpenPLC confirm only the fixed-length instance.

Counters need no clock; for CTU and CTD the tool matches the model on the fewest scans to reach the preset
($2\,\mathit{PV} - 1$ and $2\,\mathit{PV}$) in 12 of 12 cases, and MATIEC's counters agree with the model over 80,000
random scans. % paper-pin run: counters/counters.json

\subsection{The monitor fact}

A deadline is a property about history, so the per-block facts are not enough. What is enough, for one timer, is a fact
that ties the monitor to the timer: while the monitor counts, $c \le T_{\max} + \mathit{ET}$ and
$\mathit{ET} + 1 \le \mathit{PT}$, so $c \le T_{\max} + \mathit{PT} - 1$. With this fact (\path{--ld-monitor-invariants}, a prototype on the pinned build that is not yet submitted upstream)
k-induction proves the exact deadline at $k = 2$ for TON, TOF, and TP with scans of 10 to 30~ms and presets from 50~ms to
20~s, in under 0.2~s each, where \gls{bmc} timed out from 200~ms. The fact is checked like the block facts, so by
\cref{lem:inv} it adds no wrong SAFE verdict and no spurious violation to the timing-mode program. % paper-pin run: logs/scale_s4.log
For a fixed scan length the exact bound is \textit{PT}, below the fact's bound, and induction returns UNKNOWN; for those rows a
bounded \gls{bmc} run over the worst-case horizon finds no violation. The fact applies only to a single timer whose trigger
and response are its own variables or one-contact copies of them; it does not cover a cascade or the stairs-light program,
where the response combines the timer's $Q$ with two other variables.

The deadline results of RQ4 hold within these limits: the generated facts make k-induction converge on properties that follow
from them in one step, and the monitor fact makes it converge on single-timer deadlines over a scan range. \gls{bmc}
proves everything else about elapsed time up to a bound: 201 scans in 32~s for a fixed scan length, and a timeout after
300~s at 9 scans for a range (\cref{sec:deadlines}).

\section{Defects found while building the frontend}
\label{app:defects}

While we developed ESBMC-PLC, the four checks of this paper (the proof, the timing model, translation validation, and
a differential run against the K-LD executable semantics and the OpenPLC runtime) exposed defects in earlier versions of
the frontend. \Cref{tab:defects} lists them, grouped by what they do to a verdict. An \emph{unsound SAFE} hides a real
violation; a \emph{spurious alarm} reports one that no run has; a \emph{divergence} makes the model differ from MATIEC in
a way that can go either way.

\begin{table}[t]
\centering
\caption{Defects in the \gls{esbmc} \gls{ld} frontend exposed during development. Issue and pull request numbers refer to
esbmc/esbmc. ``Refused'' means the input is rejected with a diagnostic instead of modeled. $^\dagger$Fixed upstream (\iss{8859}, \iss{8907},
\iss{8910}) after the evaluated build branched from upstream \gls{esbmc} on October 12, 2026; the evaluated build, and so every
result here, does not contain these fixes.}
\label{tab:defects}
\small\begin{tabular}{@{}>{\raggedright\arraybackslash}p{0.43\linewidth}>{\raggedright\arraybackslash}p{0.13\linewidth}>{\raggedright\arraybackslash}p{0.22\linewidth}>{\raggedright\arraybackslash}p{0.14\linewidth}@{}}
\toprule
Defect & Effect & Exposed by & Status \\
\midrule
Timer skipped on a rung path, then SUCCESSFUL (stairs light) & unsound SAFE & K-LD and OpenPLC differential & fixed \\
Untranslated function-block body left its outputs stale & unsound SAFE & code audit & fixed (\iss{8079}) \\
Operator block or timer pin wired to an output dropped or read as 0 (\iss{7389}) & unsound SAFE & frontend audit & refused (\iss{8064}) \\
Variable written by another task kept as program state & unsound SAFE & I/O classification argument & fixed (\iss{8065}) \\
Non-timer block dropped from a coil with another live input & unsound SAFE & code audit, then probe & refused (\iss{8435}) \\
Seven inputs accepted and modeled as something else & unsound or wrong & code audit, then probes & refused (\iss{8521}) \\
TON output true on the rising scan for $\mathit{PT} \le 1$ (\iss{8178}) & spurious alarm & MATIEC route on a case study & fixed (\iss{8179}) \\
Timer output read only by a contact treated as an input (\iss{8444}) & spurious alarm & timing model & fixed (\iss{8445}) \\
Wired function-block inputs replaced by nondeterminism (\iss{7580}) & spurious alarm & suite runs & fixed (\iss{8034}) \\
Boolean operators dropped in function-block bodies (\iss{7579}) & divergence & suite runs & fixed (\iss{8035}) \\
CTD never reloaded (\iss{7353}) & divergence & suite runs & fixed (\iss{8033}) \\
Scan order and previous-scan reads (\iss{7352}) & divergence & suite runs & fixed (\iss{8036}) \\
Counter value ran past its preset & divergence & MATIEC library source & fixed (\iss{8066}) \\
INT modeled as 32 bits (\iss{8194}) & divergence & translation validation & fixed (\iss{8198}) \\
User function block \texttt{EN}/\texttt{ENO} ignored (\iss{8210}) & divergence & translation validation & open$^\dagger$ \\
Edge contact compares with the end-of-scan latch (\iss{8211}) & divergence & semantics against the standard & open$^\dagger$ \\
REAL modeled as a double (\iss{8212}) & divergence & translation validation & open$^\dagger$ \\
\bottomrule
\end{tabular}
\end{table}

Six of the 17 entries could make the tool prove a property that fails. All six are fixed or refused, and the
evaluation in \cref{sec:evaluation} runs on a build that contains the fixes. A further audit, with probe programs
compiled by MATIEC, found 11 more inputs the frontend accepted without modeling them: a wire cycle through a counter's
reset, a declared name equal to a monitor's or the cycle delta's, a negative preset, a preset that changes, a second writer of a
block's state, a TOF, TP, or counter that starts nonzero, an action, a power-flow wire from \textit{ET} or \textit{CV}, a
reference to a generated name, a timer and a counter sharing a name, and a non-Boolean property atom. The evaluated build
refuses all 11, and none of them occurs in the suite, so no verdict of \cref{tab:rq1} changes.
Three divergences remain open, and Appendix~\ref{app:departures} carries them, with the TP case of
\cref{sec:refinement}, as listed departures.

No single technique found most of these. Translation validation exposed the integer width and the enable pins, an
independent timing model exposed the havocked timer output, code audits followed by small probe programs
exposed the dropped block and the seven silently accepted inputs, and the differential with K-LD and OpenPLC exposed the
skipped timer. We read this as the case for running all of them: each one saw a class of defect
the others could not.

\section{Evaluation details}
\label{app:evaluation}

\subsection{Monitors and verdicts}

Besides the five property kinds of \cref{tab:monitors}, the suite uses a loop budget for the Structured Text bodies of
user function blocks, which lie outside \cref{thm:scan} (H6). Each \texttt{WHILE} loop counts its iterations within one
scan and asserts at most $B$, the overrun a \gls{plc} watchdog would trip on; the suite sets $B = 8$. A loop that never ends
violates the budget, but so does one that ends after nine iterations, so a violation is a scan overrun, not a proof of
non-termination.

A monitor never observes the power-up store, because it runs after the first scan. An output that is true at power-up
and overwritten in the first scan therefore violates no property in the model, though it would in the real power-up
image.

\subsection{Overflow runs}
\label{app:overflow}

Of the 134 SAFE verdicts, 114 come from programs with no arithmetic that could overflow, 13 from timer and counter
programs whose overflow run finds none in 20 scans, and 7 from third-party programs whose user function block computes
\texttt{IN1 - 5} on an unconstrained INT. There the overflow run finds the wrap, H1 fails, and \cref{tab:rq1} does not
count the verdict as correct. All seven, and every other variant with a user function block, sit on their own row of
the table; 16 of those 21 variants, 13 of them accepted, are checked against the loop budget rather than a safety
property.

\subsection{Harness for the programs mined from GitHub}
\label{app:harness}

The harness failed on 11 of the 115 accepted project programs of \cref{tab:tv-mined}: Beremiz or MATIEC could not load
the file (5), the reference builder has no free value for a TIME variable (4), one query ran out of its 120~s, and one
had nothing to compare. To make files load, the harness removes I/O addresses, adds a missing namespace or header,
splits \texttt{-{}-} inside comments, and renames an instance named like its type; a variable the frontend does not
read as an input, because the body never uses it or keeps its initial value as a preset, is left out of the comparison
on both sides. Both sides read the same edited file, and no edit changes what a scan computes.

\subsection{Refused programs mined from GitHub}
\label{app:refusals}

The frontend refuses 298 of the 413 project programs. The leading causes are an operator block such as ADD or MOVE on a
data wire (58), an integer or bit-string type other than INT and DINT (35), more than one task or program (26), timer or
counter wiring outside the subset (26), a second language in the project (25), a coil wired on to further elements
(20), and a negated or edge coil (19); 2 contain an \gls{ld} jump or label (Appendix~\ref{app:gaps}).

\subsection{Verdict accounting}
\label{app:accounting}

\Cref{tab:accounting} splits the verdicts of the \gls{ld} route by the kind of property checked, which reconciles the counts
of \cref{tab:rq1,tab:viac}. All 7 SAFE verdicts that rest on an overflow are on user function blocks, and 6 of them
check a loop budget, so the 127 correct SAFE verdicts all fall on variants with a safety property: 128 SAFE there,
minus 1 that rests on an overflow. Of the 17 refusals, 12 are variants of tasks the suite declares as \gls{fbd} or \gls{sfc}, which
leaves the 229 variants of \cref{tab:tv}; the other 5 are declared \gls{ld} and comprise one \gls{fbd} or \gls{sfc} body, three
comparison blocks in power flow, and one arithmetic output, so the frontend reads 224 variants (RQ5).

The count 13 recurs for three unrelated sets: the SAFE proofs that involve timer or counter state (\cref{sec:rq1},
Appendix~\ref{app:overflow}), the variants that differ from MATIEC (\cref{tab:tv}), and the accepted loop-budget variants
(\cref{tab:accounting}).

\begin{table}[t]
\centering
\small
\caption{Verdicts of the \gls{ld} route by the property a variant checks.}
\label{tab:accounting}
\begin{tabular}{@{}lrlrrrr@{}}
\toprule
Variants & Count & Engine & SAFE & \quad of which H1 violated & VIOLATION & Refused \\
\midrule
Safety property & 225 & k-induction & 128 & 1 & 83 & 14 \\
Loop budget & 16 & runner's choice & 6 & 6 & 7 & 3 \\
\midrule
All & 241 & & 134 & 7 & 90 & 17 \\
\bottomrule
\end{tabular}

% single-build-2026-10-22/routes/uniform_ld/records and viac/viac.jsonl; PROVENANCE.txt
\end{table}

\subsection{Vacuity}

For each invariant, mutual-exclusion, and response property in the suite, the check derives the conditions that must be
reachable for the property to constrain anything and asks \gls{esbmc} whether each is. Of 207 properties checked this way,
205 are non-vacuous and 2 are vacuous (1.0\%). In one, a fault latch that the property protects against is unreachable
in the bomb variant; in the other, the second pump of an alternation program never runs. Both are statements about the
programs, which the suite's authors can correct. Another 144 properties have shapes the check does not cover, mostly
reachability goals; 3 use an operator the check does not parse, and 20 belong to refused programs.

\subsection{Input classification}

Over the 224 \gls{ld} variants the frontend could read, it saw 1,133 declared variables: 636 (56.1\%) are inputs by their
interface section, 8 (0.7\%) by the use rule, none by address alone, and the remaining 489 are outputs or internal state
that the program owns. The use rule over-approximates, so it cannot make a SAFE verdict unsound, and the verdicts of
\cref{tab:rq1} do not depend on how it classifies anything other than these 8 variables.

\subsection{Second C verifiers on MATIEC's C}
\label{app:second-verifier}

We exported each harness of the C route in the form of the \gls{svcomp}, with \texttt{\_\_VERIFIER\_nondet} inputs and a call to
\texttt{reach\_error} for a failed claim, and checked that \gls{esbmc} finds the same violations in this form. 2LS 0.10.0
(commit \texttt{9a9ad377}, built for macOS) with \path{--k-induction} and a 120~s limit reported safe 117 of the 125
variants that the \gls{ld} route proves SAFE and the C route does not (\cref{tab:viac}) and crashed on the 8 with a TON or TOF, on an internal invariant of its \gls{ssa}
construction. On the 83 harnesses whose violation the \gls{ld} route finds, it reported 70 violations and 2 crashes, but
called 11 safe; in 10 of these \gls{esbmc} finds the violation in the same file, so its SAFE verdicts cannot be used.
We tried CPAchecker 4.2.2 with its k-induction configuration on three harnesses: a safe Boolean one, a Boolean one with a
known violation, and a TON. It found no target location in any of them, because its C frontend does not support
\texttt{\_\_builtin\_va\_arg}, which MATIEC's variadic standard functions use. With the functions that \texttt{main}
cannot reach removed, the two Boolean harnesses parsed, but both k-induction and bounded model checking used up 120~s of
CPU time in their first satisfiability check, also on the one whose violation \gls{esbmc} finds in about one second, and the
TON harness still calls the variadic \texttt{\_\_\_LE\_TIME}. We did not run it on the rest. CPAchecker ran with the arm64 Z3 of our toolchain in place of
the x86-64 Z3 that the release ships, which may contribute.

\section{Correspondence with Ebnenasir's report}
\label{app:ebnenasir}

Ebnenasir's technical report~\cite{Ebnenasir2023tr} is the closest prior formalization of ladder logic scans. It gives Promela code templates for a subset of ladder logic, not inference rules, and it lists a
mechanical proof of its own formalization as future work. Its instruction set (PLS and PLF, pulses on a rising and a falling edge; SET; RST, reset; OUT; and a count-down timer) resembles
Mitsubishi's, although the report names no vendor. \Cref{tab:ebnenasir} compares it with our
semantics construct by construct.

\begin{table}[t]
\centering
\small
\caption{Ebnenasir's report against the semantics of \cref{sec:correctness}. ``Same'' means the two agree in
substance; page numbers refer to the report's printed pages.}
\label{tab:ebnenasir}
\begin{tabularx}{\linewidth}{@{}>{\raggedright\arraybackslash}p{0.3\linewidth}>{\raggedright\arraybackslash}p{0.25\linewidth}>{\raggedright\arraybackslash}X@{}}
\toprule
Construct & His report & Relation to ours \\
\midrule
Scan cycle with latched inputs & \S2.1, p.~3 & same structure; ours is a relation over a store \\
Outputs at the end of the cycle & \S2.1, p.~3 & ours writes the store at the coil; his appendix model does the same \\
Previous-scan copies for edge tests & \S2.2, p.~4; appendix & same idea; ours keeps them only for sensed variables \\
Rungs top-down, elements left to right & \S2.1, p.~3; \S4, p.~8 & same order; ours reads a coil's contacts at the coil \\
Normally open and closed contacts & implicit in guards, pp.~6--7 & same \\
Parallel branches & guard disjunction, appendix & same for textual rungs; ours adds graphical networks \\
SET and RST & \S3.2, p.~7; appendix & same as his level-based appendix form \\
Output coil & \S3.2, p.~6 & same as his appendix; his Listing~2 is edge-guarded \\
Edge contacts \texttt{[P]}, \texttt{[N]} & none (PLS and PLF instructions instead) & not in his report \\
On-delay timer & \S3.2, pp.~7--8; Listing~5 & same observable delay by our hand analysis; his counts down \\
Time as scan counts & \S2.2, p.~4 (one example) & same model; ours gives the conversion explicitly \\
TOF, TP & ``we present only'' the on-delay timer, p.~7 & not in his report \\
CTU, CTD & counters in general only, pp.~2--3 & not in his report \\
Arithmetic and data types & none & not in his report \\
Environment & one scripted input sequence, \S5 & ours admits every input sequence \\
Multiple prioritized tasks & outline only, Listing~1 & ours refuses them \\
Timing charts, within-scan properties & \S2.2, \S4 & not in ours: our properties are checked at scan boundaries \\
Fault models and fault tolerance & \S6 & not in ours \\
Graphical networks, type rules, translation & none & not in his report \\
\bottomrule
\end{tabularx}
\end{table}

The report supports a narrow claim, and we make only that one: our scan structure (latched inputs, top-down and
left-to-right rung order, previous-value copies for edge tests) and our scan-count view of on-delay timing follow the
model of its Sections~2 and~3 and its appendix. It does not ground our rules for TOF, TP, CTU, CTD, arithmetic, edge
contacts, or graphical networks, and we do not cite it for them. The timer row rests on our reading of the report's listings; we
did not run its Promela model.

\section{Correspondence with IEC 61131-3:2025}
\label{app:iec2025}

The fourth edition of IEC~61131-3~\cite{iec61131-3-2025} replaced the third in 2025. We read its clauses on tasks,
data types, arithmetic, standard function blocks, and ladder diagrams against the semantics of
\cref{sec:correctness}, the translation of \cref{sec:approach}, and the evaluation of
\cref{sec:evaluation}. MATIEC, the reference of translation validation and of \cref{thm:refine}, implements the
2013 edition. Annex~B of the 2025 edition lists the changes between the two. For the clauses below it lists new
numbering, a new drawing of MOVE (Table~32), and a rule that a task runs its program instances in the order of their
declaration (\S6.8.2), which a subset with one program does not exercise. Its other changes (UTF-8 strings,
assertions, properties, synchronization, and the deletion of Instruction List) concern constructs outside the subset.

\Cref{tab:iec2025} gives the result clause by clause. ``Follows'' means the semantics gives the construct the meaning
the clause states. ``Departs'' means it does not, and Appendix~\ref{app:departures} says how. ``Refused'' means the
frontend rejects the construct with a diagnostic, so no verdict is given. ``Not refused'' marks the one construct the
frontend neither models nor rejects.

\begin{table}[t]
\centering
\small
\caption{The semantics and the frontend against IEC 61131-3:2025. Clause, table, and figure numbers are those of the
2025 edition.}
\label{tab:iec2025}
\footnotesize\begin{tabular}{@{}p{0.15\linewidth}p{0.33\linewidth}p{0.31\linewidth}p{0.1\linewidth}@{}}
\toprule
Clause & The standard & ESBMC-PLC & Fit \\
\midrule
\S6.8.2 Tasks & a task runs its programs periodically; no input image or output latch is defined & one periodic
task; read, execute, latch (\cref{sec:semantics}); several tasks refused & follows \\
\S8.2.2, \S8.2.3, Table 73 & the left rail is ON; a horizontal link passes its state on; a vertical link is the OR of
its left links & \cref{eq:pf}, with the left rail true & follows \\
\S8.2.7 & networks run top to bottom & textual rungs in file order; graphical coils by drawing position
(\cref{sec:graph}) & follows \\
\S8.1.5.2 & feedback through a variable is allowed; explicit loops only in \gls{fbd} & feedback through variables under
sequential evaluation; wire cycles refused & follows \\
Table 74, normally open (NO) and normally closed (NC) contacts & AND of the left link with the variable or its negation; the variable is not modified &
\cref{eq:rung}, read at the coil & follows \\
Table 74, P and N contacts & ON when a transition is sensed between evaluations of the element & compared with the
value latched at the end of the previous scan & departs \\
Table 75, coil, set, reset & store the left link, latch, or unlatch the variable & as stated; the later write wins &
follows \\
Table 75, negated and transition coils & store the inverse or an edge of the left link & not modeled & refused \\
\S8.2.5 & a coil passes the left link to its right & not modeled & refused \\
\S8.2.6 & blocks in \gls{ld} carry power flow through a Boolean input and output & TON, TOF, TP, CTU, CTD in power flow;
other blocks refused; a user block in power flow is refused (Appendix~\ref{app:gaps}) &
follows \\
\S6.6.1.5, Table 18 & a block whose \texttt{EN} is false is not executed and its \texttt{ENO} is reset; its outputs are then implementer specific &
\texttt{EN} of a user block ignored; wired \texttt{EN} or \texttt{ENO} on a standard block refused & departs \\
\S8.1.6, Table 72 & jumps, labels, and returns transfer control & refused in graphical networks, as upstream since commit \anonhash{ddae49e1a5}; dropped beside textual rungs (Appendix~\ref{app:gaps}) & refused \\
\S6.6.3.5.3, Table 45 & R\_TRIG and F\_TRIG blocks & not modeled & refused \\
\S6.6.3.5.4, Table 46 & CTU counts while $\mathit{CV} < \mathit{PV}_{\max}$, CTD while $\mathit{CV} >
\mathit{PV}_{\min}$; $Q$ compares \textit{CV} with \textit{PV} or 0 & counts stop at \textit{PV} and 0, as in MATIEC;
$Q$, reset, and load as stated & departs \\
\S6.6.3.5.5, Table 47, Figure 17 & TON and TOF timing; TP pulse, then \textit{ET} back to 0 once \textit{IN} is low &
TON and TOF at scan boundaries under P1, P2, P4 (\cref{thm:refine}); TP \textit{ET} held at \textit{PT} & follows;
TP departs \\
Table 47, note & a change of \textit{PT} while timing is implementer specific & a preset that can change is refused
(P4) & refused \\
\S6.4.2, Table 10 & INT 16 bits, DINT 32 bits, REAL single precision (ISO/IEC 60559), TIME range implementer
specific & INT 16, DINT 32, TIME 32 bits in ms; REAL double; unsigned, 8- and 64-bit types refused & follows; REAL
departs \\
\S6.6.2.5.9 & a result outside its type, or a division by zero, is an error; detection is implementer specific &
division by zero asserted on every run (H2); overflow checked by a bounded run for each SAFE verdict (H1); the
semantics wraps & follows \\
\bottomrule
\end{tabular}
\end{table}

\paragraph{What the departures mean for the results}
\Cref{thm:scan} relates the GOTO program to the semantics, so no departure affects it; a departure limits what a verdict
says about a controller that follows the standard. On the suite it limits little. No \gls{ld} variant uses an edge contact
(the running example uses \texttt{-[P]-} but is not in the suite, and the R\_TRIG and F\_TRIG of \cref{tab:plcverif}
are blocks, not contacts) or REAL, no property reads a counter's \textit{CV} or a timer's \textit{ET}, and no program contains a jump; the
departures that remain apply to the 21 variants with a user function block, which \cref{tab:rq1} reports apart and
which cause every difference of \cref{tab:tv}. For a program outside the suite, a verdict carries over to a
conforming controller when the program avoids the constructs marked ``departs'' and the one marked ``not
refused''.

\end{document}